\pdfoutput=1
\documentclass[runningheads]{llncs}
\usepackage{graphicx}
\usepackage{grffile}
\usepackage{comment}
\usepackage{tabularx}

\usepackage{booktabs}
\usepackage{siunitx}
\usepackage{amsmath,amssymb} 
\usepackage{color}
\usepackage{soul}
\usepackage[ruled,vlined]{algorithm2e}

\usepackage[super]{nth}

\usepackage{hyperref}
\hypersetup{
    colorlinks=true,
    linkcolor=blue,
    filecolor=magenta,      
    urlcolor=cyan,
}
\usepackage{cleveref}
\crefname{section}{Sec.}{Sections}
\crefname{figure}{Fig.}{Figures}
\crefname{table}{Tab.}{Tables}
\crefname{equation}{Eq.}{Equations}
\crefname{appsec}{Appendix}{Appendices}

\usepackage{subcaption}
\usepackage{soul}
\makeatletter
\DeclareRobustCommand\onedot{\futurelet\@let@token\@onedot}
\def\@onedot{\ifx\@let@token.\else.\null\fi\xspace}
\makeatother

\newcommand{\cf}{cf\onedot}
\newcommand{\ie}{i.\,e.,\xspace}

\makeatletter
\newcommand{\printfnsymbol}[1]{%
  \textsuperscript{\@fnsymbol{#1}}%
}
\makeatother

\begin{document}

\pagestyle{headings}
\mainmatter
\def\ECCVSubNumber{14}  

\title{Understanding Compositional Structures in Art Historical Images using Pose and Gaze Priors\\
\small \emph{Towards Scene Understanding in Digital Art History}}

\titlerunning{Image Compositions in Art History}

\author{Prathmesh Madhu\inst{1}\orcidID{0000-0003-2707-415X}\thanks{equal contribution, [$*$] - ORCiD IDs} \and
Tilman Marquart\inst{1}\printfnsymbol{1} \and
Ronak Kosti\inst{1}\orcidID{0000-0003-2453-7876} \and 
Peter Bell\inst{2}\orcidID{0000-0003-4415-7408}
\and Andreas Maier\inst{1}\orcidID{0000-0002-9550-5284} \and 
Vincent Christlein\inst{1}\orcidID{0000-0003-0455-3799}}

\authorrunning{Madhu and Marquart et al.}

\institute{Pattern Recognition Lab, \url{https://lme.tf.fau.de/} \and Institute for Art History, \url{https://www.kunstgeschichte.phil.fau.de/} Friedrich-Alexander-Universit\"at Erlangen-N\"urnberg, Germany\\
Email: \email{prathmesh.madhu@fau.de}
}
\maketitle
\begin{abstract}\footnote{The final authenticated version will be available: \href{https://link.springer.com/conference/eccv}{ECCV 2020 Workshops Link}}
Image compositions as a tool for analysis of artworks is of extreme significance for art historians. These compositions are useful in analyzing the interactions in an image to study artists and their artworks. Max Imdahl in his work called \emph{Ikonik}, along with other prominent art historians of the \nth{20} century, underlined the aesthetic and semantic importance of the structural composition of an image. Understanding underlying compositional structures within images is challenging and a time consuming task.  Generating these structures automatically using computer vision techniques (1) can help art historians towards their sophisticated analysis by saving lot of time; providing an overview and access to huge image repositories and (2) also provide an important step towards an understanding of man made imagery by machines. In this work, we attempt to automate this process using the existing state of the art machine learning techniques, without involving any form of training. Our approach, inspired by Max Imdahl's pioneering work, focuses on two central themes of image composition: (a)~detection of action regions and action lines of the artwork; and (b)~pose-based segmentation of foreground and background. Currently, our approach works for artworks comprising of protagonists (persons) in an image. In order to validate our approach qualitatively and quantitatively, we conduct a user study involving experts and non-experts. The outcome of the study highly correlates with our approach and also demonstrates its domain-agnostic capability. We have open-sourced the code: \href{https://github.com/image-compostion-canvas-group/image-compostion-canvas}{https://github.com/image-compostion-canvas-group/image-compostion-canvas}

\end{abstract}

\section{Introduction}
Understanding narratives present in an artwork has always been a challenge and is strongly researched since the late \nth{19} century~\cite{artcv}. A high-level interpretation for the given scene is more ambiguous than a low-level interpretation~\cite{bienenstock1997compositionality} meaning it is easy to recognize and interpret small objects and characters in the image rather than presenting a high level abstract description of any scene. In the recent times of deep learning, numerous highly  successful supervised techniques have been presented for various applications like object detection~\cite{Gonthier18}, person identification~\cite{artchars}, segmentation~\cite{ronneberger2015u}, retrieval~\cite{jenicek2019linking} etc. However, these methods have some major drawbacks. First, even though they perform extremely well on the benchmark datasets, they fail to generalize across other real-world datasets. The reason being that it is extremely difficult to capture the real world distribution and it's complexity within a single large dataset. Second, these deep models are very sensitive and error-prone compared to humans who naturally adapt to the visual context as stated in~\cite{deepnetsyuille2018}.

One solution suggested in the DeepNets~\cite{deepnetsyuille2018} paper is the idea of using the compositionality of the images to generate models that can be trained on finite datasets, and then can be generalized across unseen datasets. The assumption is that the structures within an image are composed of various substructures following a grammatical set of rules, which can be learned from a finitely annotated datasets. These compositional models can be used to reason and diagnose the system, extrapolate beyond data and answer varied questions based on learned knowledge structure. On a parallel front, from a theoretical perspective, Bienenstock~\emph{et al.}~\cite{bienenstock1997compositionality} suggested that the hierarchical compositional structure of natural visual scenes can be reduced down to a collection of drastic combinatorial restrictions; meaning one can break down any scene into 2D projections of objects present in the scene. From this perspective, it can be understood that \emph{composition} is one of the fundamental aspects of human cognition. 

\begin{figure}[t]
		\centering
		\begin{subfigure}[t]{0.25\linewidth}
			\centering
			\includegraphics[width=0.9\linewidth, keepaspectratio]{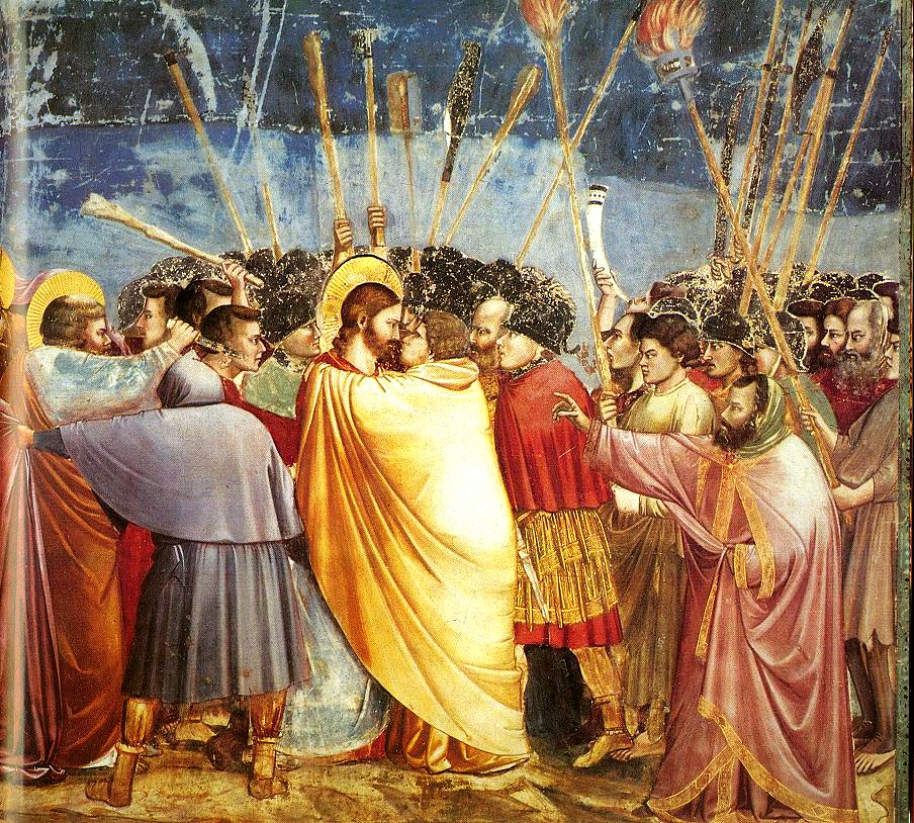}
			\label{fig:main-a}
		\end{subfigure}%
		\begin{subfigure}[t]{0.25\linewidth}
			\centering
			\includegraphics[width=0.9\linewidth, height=3cm, keepaspectratio]{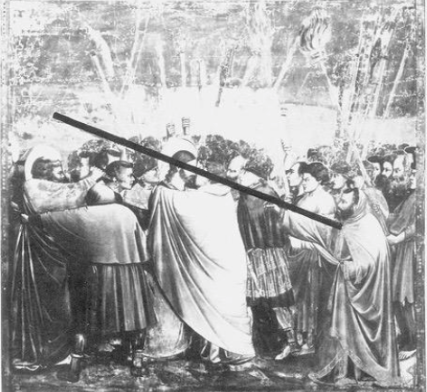}			
			\label{fig:main-b}
		\end{subfigure}%
		\begin{subfigure}[t]{0.25\linewidth}
			\centering
			\includegraphics[width=0.9\linewidth, height=3cm, keepaspectratio]{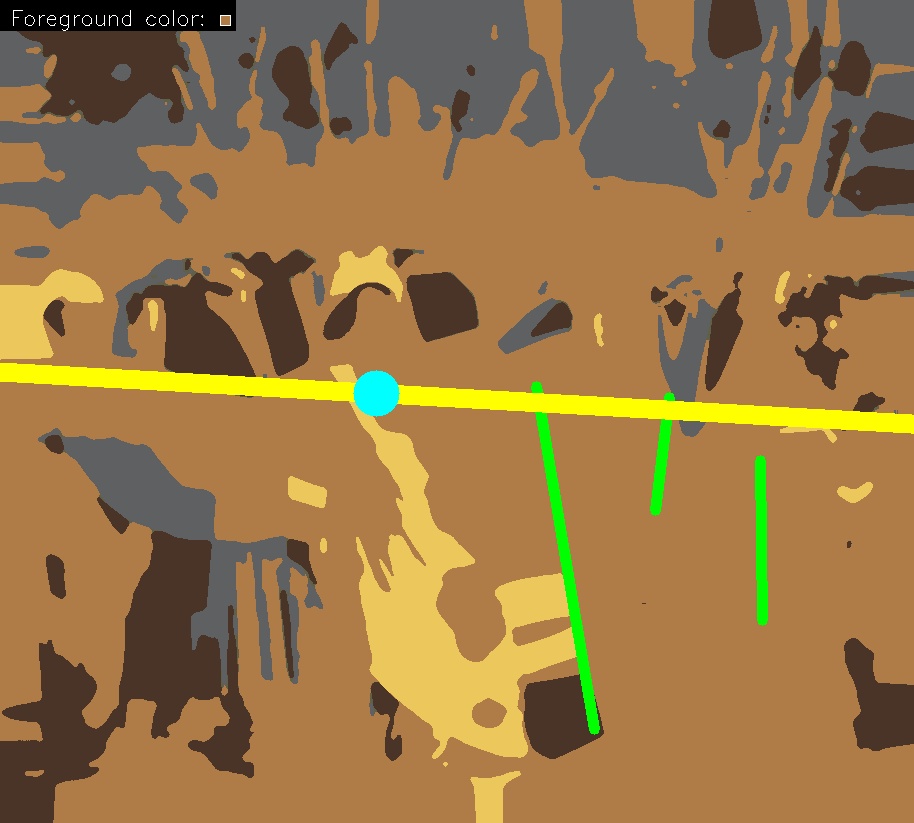}
			\label{fig:main-c}
		\end{subfigure}%
		\begin{subfigure}[t]{0.25\linewidth}
			\centering
			\includegraphics[width=0.9\linewidth, height=3cm, keepaspectratio ]{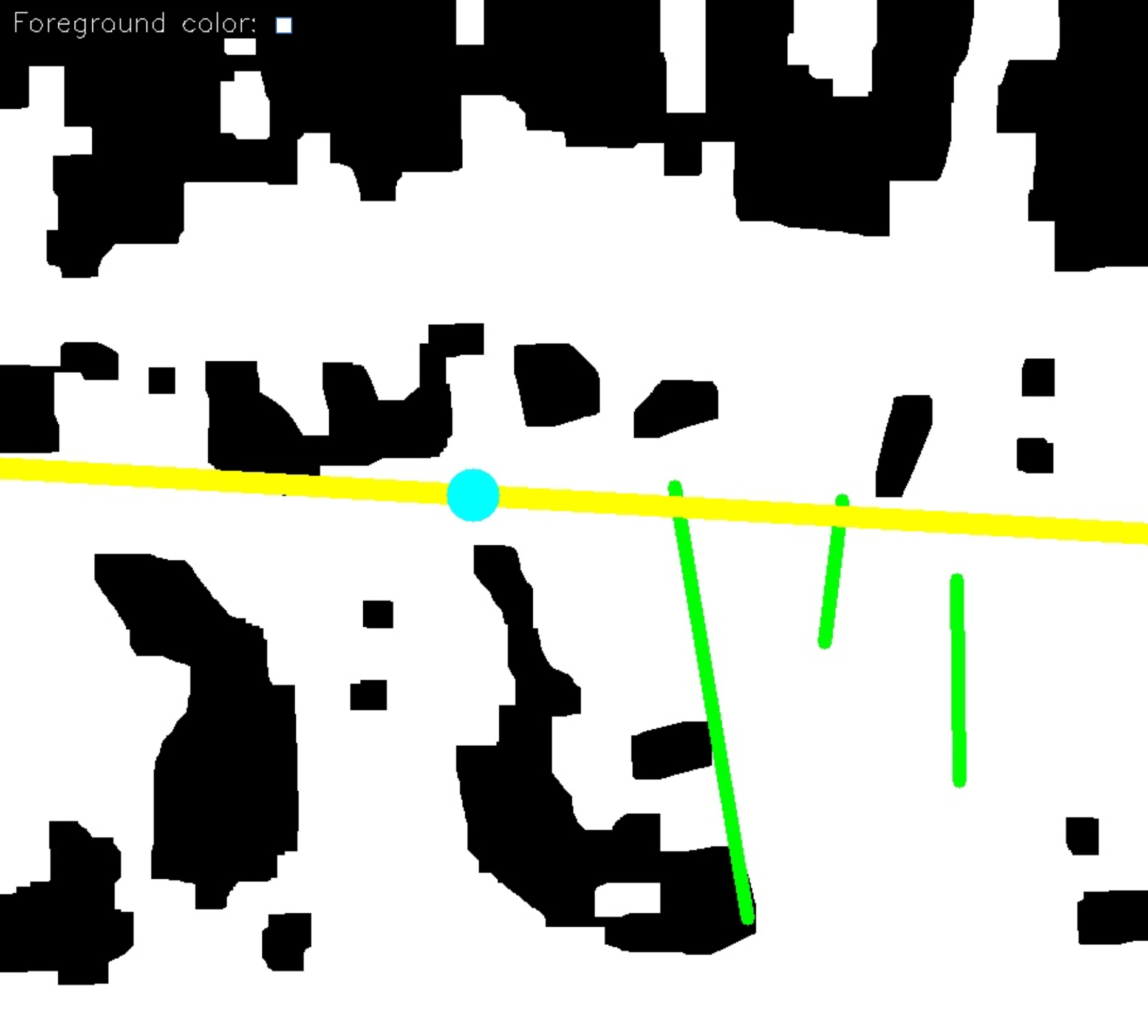}
			\label{fig:main-d}
		\end{subfigure}
		\begin{subfigure}[t]{0.25\linewidth}
			\centering
			\includegraphics[width=0.9\linewidth, height=3cm, keepaspectratio]{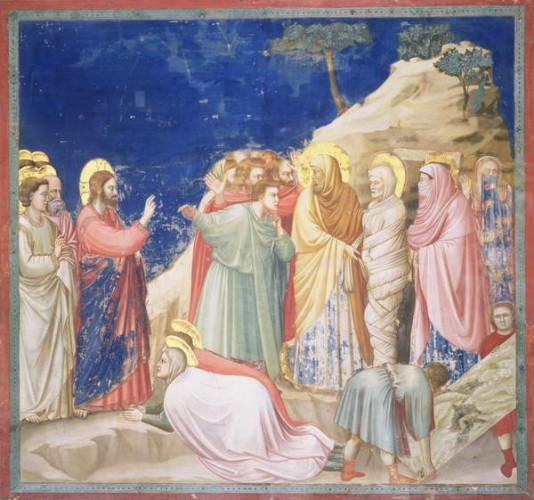}
			\caption{\textbf{Original Art}}
			\label{fig:main-e}
		\end{subfigure}%
		\begin{subfigure}[t]{0.25\linewidth}
			\centering
			\includegraphics[width=0.9\linewidth, height=3cm, keepaspectratio]{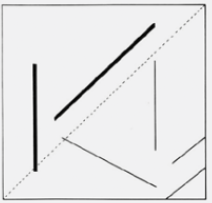}
			\caption{\textbf{Imdahl's Take}}
			\label{fig:main-f}
		\end{subfigure}%
		\begin{subfigure}[t]{0.25\linewidth}
			\centering
			\includegraphics[width=0.9\linewidth, height=3cm, keepaspectratio]{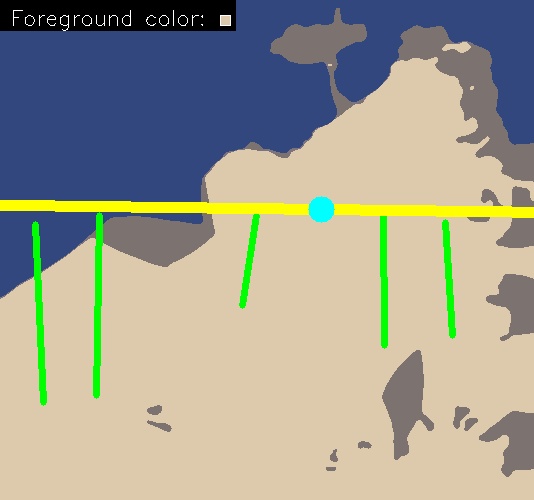}
			\caption{\textbf{ICC}}
			\label{fig:main-g}
		\end{subfigure}%
		\begin{subfigure}[t]{0.25\linewidth}
			\centering
			\includegraphics[width=0.9\linewidth, height=3cm, keepaspectratio]{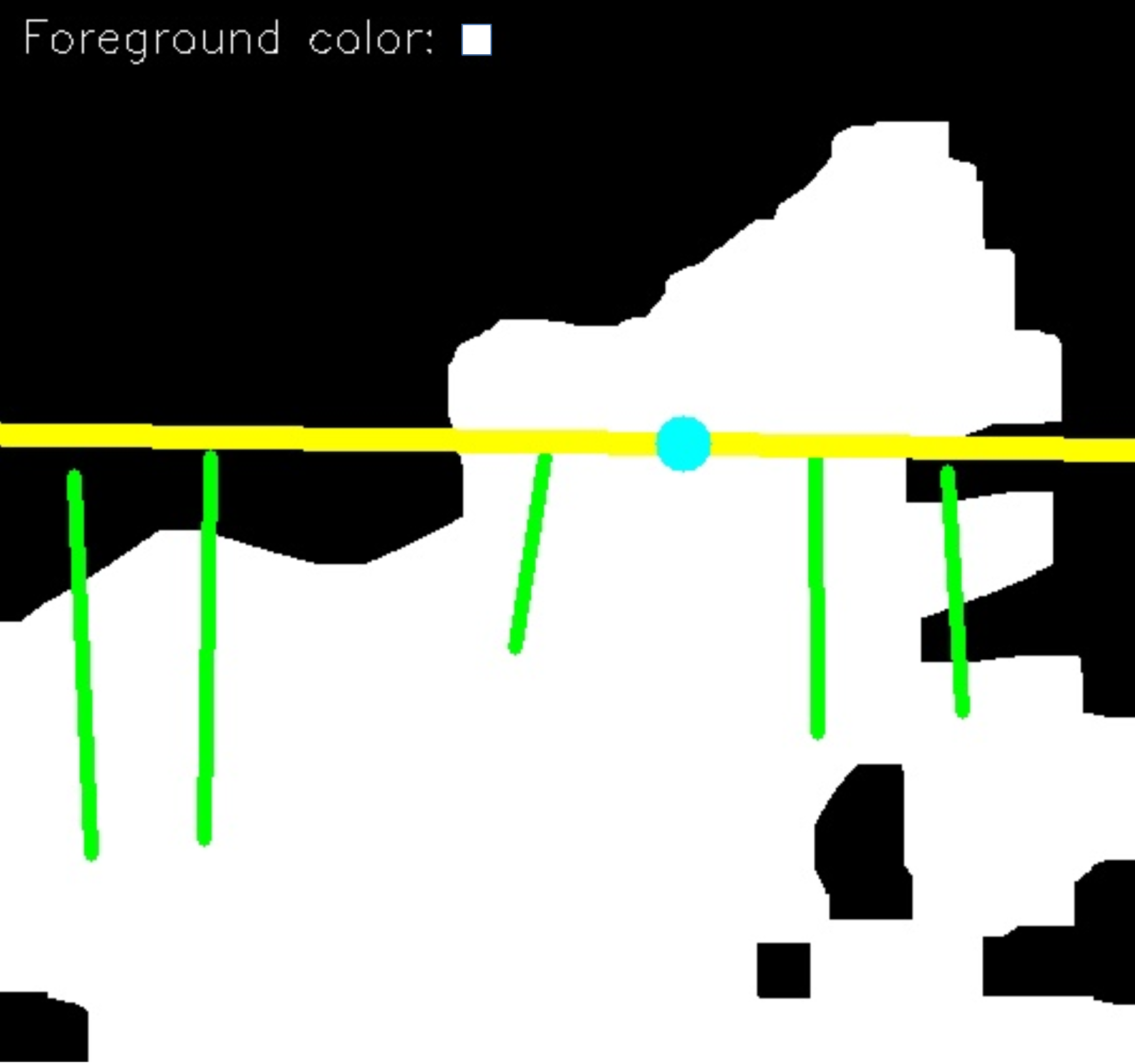}
			\caption{\textbf{FG/BG Sep.}}
			\label{fig:main-h}
		\end{subfigure}
    \caption{\protect{\subref{fig:main-e}} Giotto's Original Paintings, (top) ``The kiss of Judas", (bottom) ``Raising of Lazarus"; \protect{\subref{fig:main-f}} Imdahl's compositional analysis using structural elements for Giotto's works; \protect{\subref{fig:main-g}} Their corresponding Image Composition Canvas (ICC); \protect{\subref{fig:main-h}} Binarized ICC to highlight the foreground/background separation.}
    \label{fig:main}
	\end{figure}

Motivated by this understanding of compositional structures within images, in this paper, we investigate the problem of determining and analyzing the composition of various scenes in art history. Specifically motivating is the work of Max Imdahl called \emph{Ikonik}~\cite{imdahl1975giotto}, where he formulates a methodology using an artwork's structure to determine its significance. His work is considered as a model of image analysis, which can be considered as a complement to Panofsky's iconology~\cite{ionescu2014you}. He argues that visual cues or the ``visual seeing'' overpowers biblical or textural references by giving references to artworks by Giotto, \cf \cref{fig:main}. Giving the examples of \emph{Ascent to Cavalry} and \emph{The kiss of Judas} (\cref{fig:main-e}), he explains how the structural relation between foreground and background, the distribution of colors and dynamism between the characters can help in understanding artists and their artworks. 

Motivated by his work, we propose an non-supervised computational approach to find compositional structures in an image. Our algorithm is driven by the human perception that posture and gaze of the main protagonists help to identify the region of interest (action regions) of any scene (\cref{fig:main-g,fig:main-h}). We enable pre-trained OpenPose~\cite{caoOpenPoseRealtimeMultiPerson2019} framework to detect pose-keypoints of protagonists and further exploit them to get a pseudo gaze-estimate without using any existing state of the art gaze detection methods which would have required largely annotated data. We combine all these compositional elements and plot it on an empty canvas which gives an estimate of the representation of the underlying structure and composition within a given scene, which we call \emph{Image Composition Canvas} (ICC). These compositional elements of ICC can later be used as visual features to cross-retrieve images from various datasets. 
Our method generates \emph{ICCs} that contains two important aspects of image composition: (1)~constructing the global action lines and action regions (2)~pose-based semantic separation of foreground and background. The detected pose-keypoints~\cite{caoOpenPoseRealtimeMultiPerson2019} not only assist in foreground/background extraction, but also help in generating the global action lines and action regions. The result is a ``visual seeing'' which concentrates strictly on the compositional structure of an image and is, thereby, complementary to human perception usually driven by the semantics of the narration. 


\section{Related Work}\label{ReWo}
\noindent \textbf{Semblances in Art History}
Max Imdahl showed that discovering similar painting structures or compositional elements is an important aspect in the analysis of artworks~\cite{imdahlGiottoArenafreskenIkonographie1980}. However, detecting the relevant features for the same is a relatively novel task for computer vision. Previous works mainly focus on features for image retrieval, for example, in \emph{SemArt}~\cite{garcia2018read} the paintings are aligned with its attributes such as title, author, type, time and captions by a neural network which learns a common embedding space between them. In another example, these attributes (also called contextual information) are trained jointly in a multi-task manner to achieve a context-aware embedding~\cite{garcia2019context}, thereby improving the retrieval performance through a knowledge graph that is created using attributes as prior information. In \emph{Artpedia}~\cite{artpedia2019}, the authors are able to align visual and textual content of the artwork without paired supervision. Jenicek~\emph{et al.} come closest to our approach in that they link paintings through a two step retrieval process, by finding similarity-of-pose across different motives~\cite{jenicek2019linking,bell_ikonographie_2019}. They also explore the underlying visual correlation between artworks.\\

\noindent \textbf{Human Pose and Gaze Estimation}
Human posture is a strong marker for compositional element of the image. It helps to understand the visual relationships depicted in the image. For the semantic understanding of a scene, it becomes important to analyze the poses of all persons in the given scene. Multi-Person Pose Estimation (MPPE) is a challenging task due to occlusions, varying interactions between different people and objects. MPPE could be divided into \emph{top-down} (detect person $\to$ predict pose point for each)~\cite{papandreou2017towards} and \emph{bottom-up} (detect body joints for all persons $\to$ join the points to detect poses)~\cite{caoOpenPoseRealtimeMultiPerson2019} approaches. The current state-of-art method uses \emph{PoseRefiner}~\cite{Fieraru2018CVPRW} as a post-processing technique to refine the pose estimation to get the best results on the MPII dataset~\cite{andriluka14cvpr}. Empirical tests showed OpenPose~\cite{caoOpenPoseRealtimeMultiPerson2019} performed well on our art-historical dataset, so we use it for all our experiments. 

Importance of human gaze is evident in the related work of detecting human gazes for automatic driver analysis and understanding attention spans. Gaze360~\cite{kellnhofer2019gaze360} works on a diverse set of environments, and it is well suited for video or multi-frame inputs depicting temporal relationship which is absent in our data. Another method, \textit{Where are they looking?}~\cite{recasensWhereAreThey2015}, uses head location and the image to predict the gaze direction in the image. This method requires gaze annotations (eye locations and gaze directions), however, since our data is a specialized collection of images chosen to study the compositions in a non-supervised manner, this approach could not be applied for our work. Hence, we derive gaze-bisection-vectors from pose keypoints for gaze priors. \\

\noindent \textbf{Foreground/Background Separation.}
Detecting foreground and background has been one of the important pre-processing task for computer vision before the deep learning techniques came into practice. Separating foreground from background helps to focus on where the object/region of interest lies at. For example, in image-reconstruction-guided landmark detection, the algorithms are quite often assisted by fine-grained separation of background and foreground techniques~\cite{dundar2020unsupervised,lorenz2019unsupervised,jakab2018unsupervised}. Specifically,~\cite{dundar2020unsupervised} factorizes the reconstruction task to achieve better landmark-detection for foreground objects by simultaneously improving the background rendering. 
    
Exploiting this prior information about the separation of foreground and background is very useful even for existing state of the art methods. For example, in single stage object detectors the high number of candidate locations ($\approx$100\,k) creates huge bias towards background classes~\cite{liu2016ssd}. Dynamically weighted focal loss~\cite{lin2017focal} mitigates this imbalance by heavily penalizing the learning model for incorrect foreground objects, thereby forcing the model to attend more to the foreground objects. Our approach inherently uses pose estimates of the protagonists. We use an enhanced k-means clustering with the pose estimates to achieve foreground background separation.

Our algorithm is motivated by how a human understands the composition of scenes in paintings. In brief, our main contributions include: (1)~Generation of Image Composition Canvas (ICC) which showcases the global and local action lines and action regions, (2)~Semantic separation of foreground/background, (3)~Generizability of the approach to images leading a step towards domain-agnostic modeling.    
    
\begin{figure}[t]
	\centering
	\includegraphics[width=0.85\linewidth, height=0.45\linewidth]{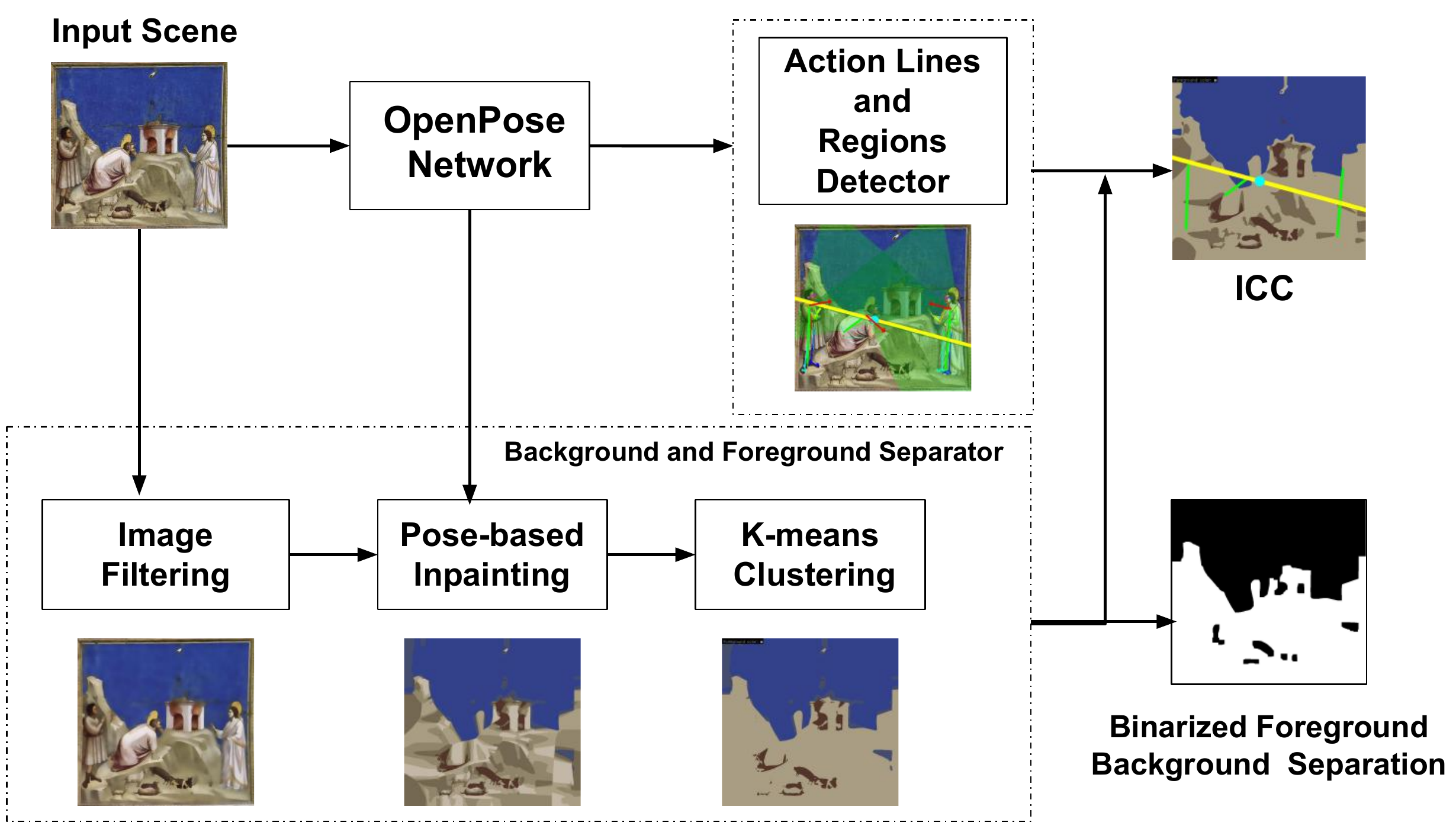}
	\caption{Generating Image Composition Canvas (\emph{ICC}): Proposed \emph{ICC} pipeline along with foreground/background separation from images. First row details in~\protect{\cref{fig:pose}}}
	\label{fig:pipeline}
\end{figure}	

\section{Methodology}\label{Meth}
We propose a non-supervised approach to generate \emph{ICC} to assist our understanding of underlying structures in art historical images. Our approach uses a pre-trained OpenPose network~\cite{caoOpenPoseRealtimeMultiPerson2019}, image processing techniques and a modified k-means clustering method. The pipeline of the proposed algorithm is shown in \cref{fig:pipeline} (Fig.~\ref{fig:pose} shows the visual counterpart). Our method consists of two main branches: (1)~a detector for action lines and action regions (\cref{sec:actions}) and (2)~the foreground/background separator (\cref{sec:backfore}). We use the estimated poses to detect pose triangles 
and propose a simple technique which we call gaze cones to obtain gaze directions estimates 
without involving any training or fine-tuning. 
We also use the detected keypoints for foreground/background separation (\cref{sec:backfore}). We combine this information and draw a final ICC that estimates the image composition of the given scene under study (\cref{sec:icc}). We evaluate the proposed method by performing a user-study where we ask domain experts and non-experts to annotate action lines (global and local) and action regions, details of which are mentioned in \cref{sec:userstudy}
    
\subsection{Data Description}\label{Data}
 The dataset contains $20$ images of fresco paintings from the $13th$ century. \cref{fig:main-e,fig:circlemap-b,fig:fgbg-a} show some examples of Giotto's famous fresco paintings. The paintings come from the Scrovegni chapel also known as arena chapel in Padua, Italy and were made by the painter Giotto di Bondone (considered the decisive pioneer of the Italian Renaissance). Most of these images have been analyzed by Max Imdahl in his book \emph{Ikonik Arenafresken}~\cite{imdahlGiottoArenafreskenIkonographie1980}. We also use 10 random images containing people from COCO test dataset, 5 ``Annunciation of Our Lady" and 5 ``Baptism of Christ" images for evaluation and to show that our approach is domain-agnostic. 
    
\subsection{Action Lines and Action Regions}\label{sec:actions}
Global \textit{Action line} (AL) is the line that passes through the main \emph{activity} in the scene, which normally is also aligned with the central protagonists. Local \textit{Action Line} or \textit{Pose Line} (PL) represent the poses of the protagonists, abstracted in the form of a line. \textit{Action region}(s) (AR) is(are) the main \emph{region(s) of interest}, more often than not it is the region where gazes of all the characters theoretically meet, or the focus of their attention. 
    
\begin{figure}[t]
	\centering
	\begin{subfigure}[t]{0.11\linewidth}
		\centering
		\includegraphics[height=1.7\linewidth]{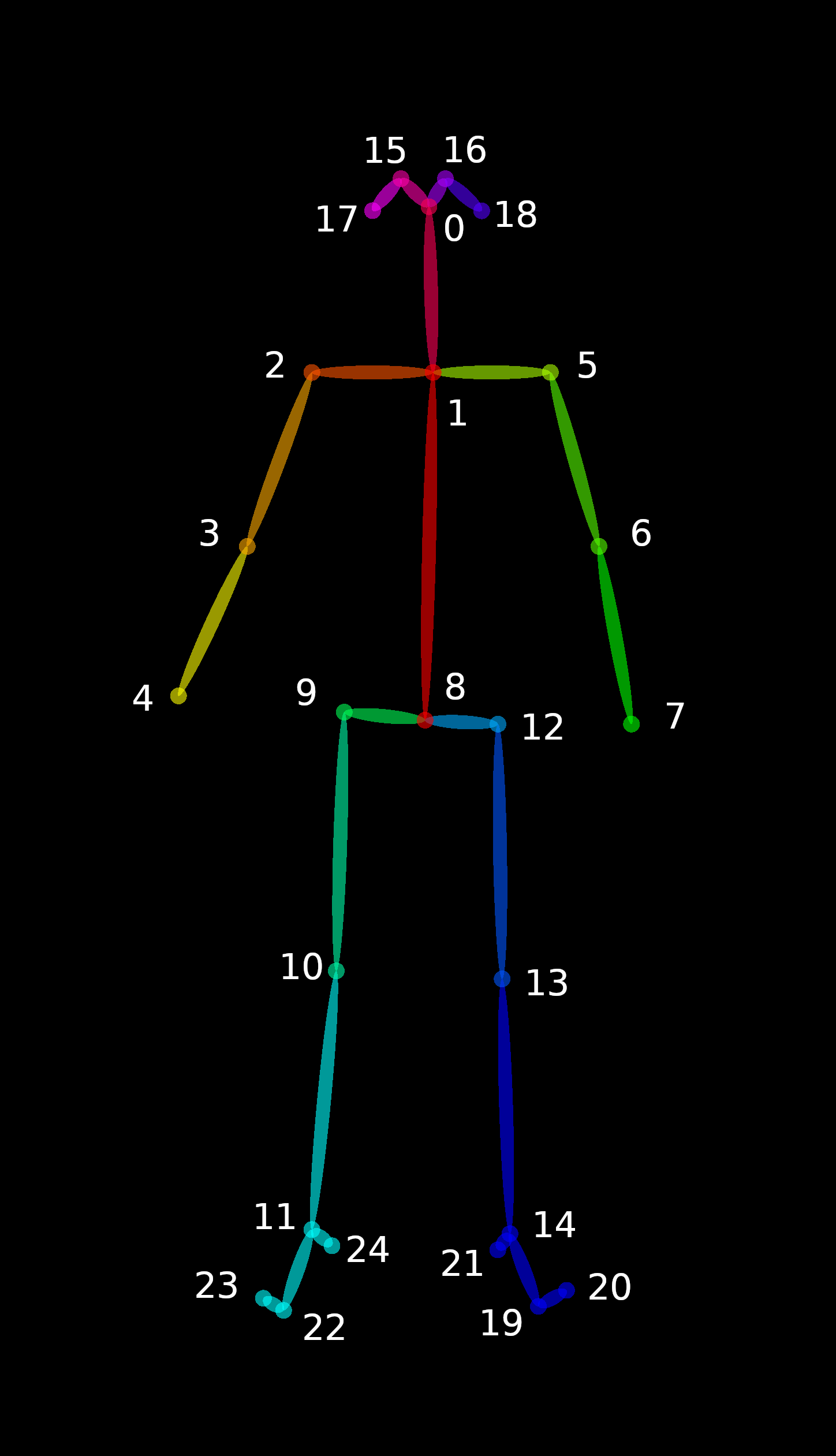}
		\caption{}
		\label{fig:pose-d}
	\end{subfigure}%
	\begin{subfigure}[t]{0.11\linewidth}
		\centering
		\includegraphics[height=1.7\linewidth]{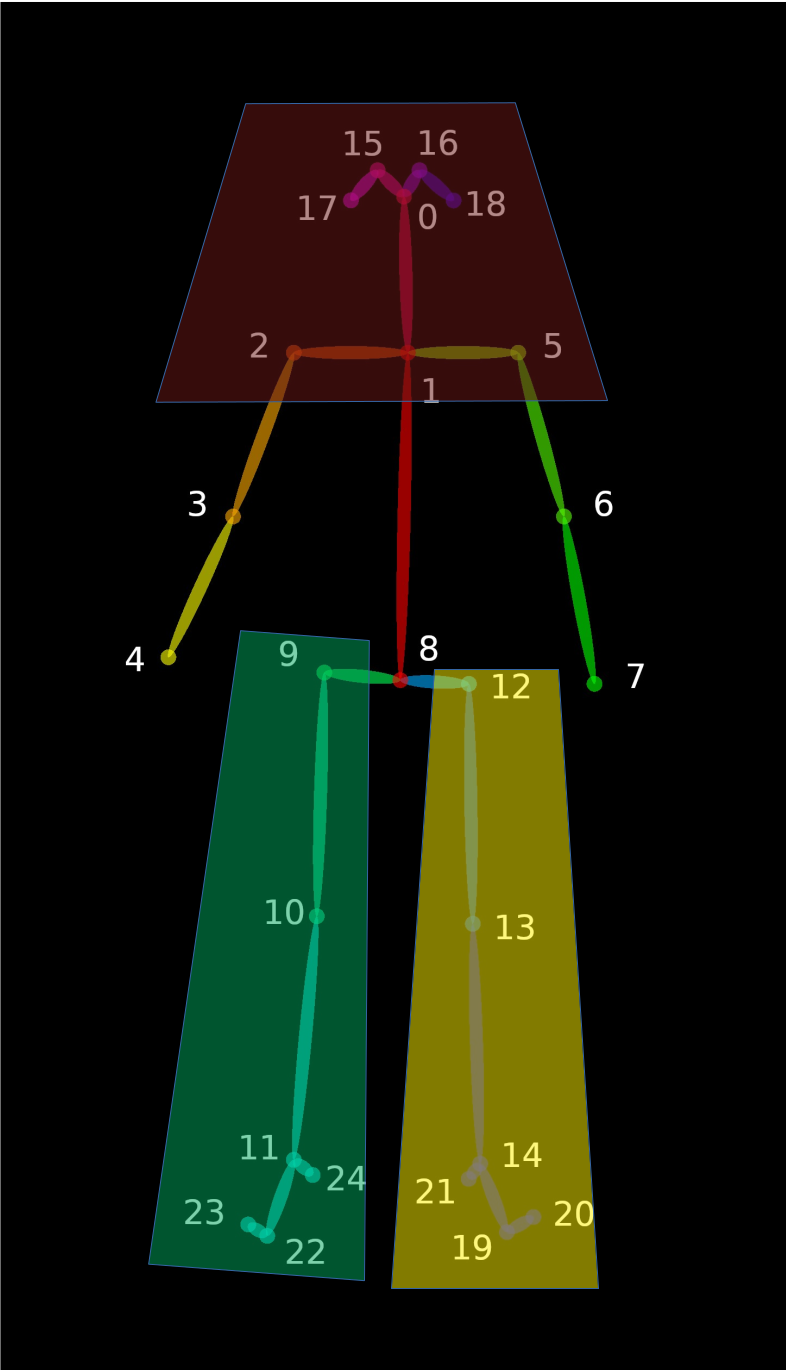}
		\caption{}
		\label{fig:pose-e}
	\end{subfigure}%
	\begin{subfigure}[t]{0.25\linewidth}
		\centering
		\includegraphics[width=0.9\linewidth, height=0.75\linewidth]{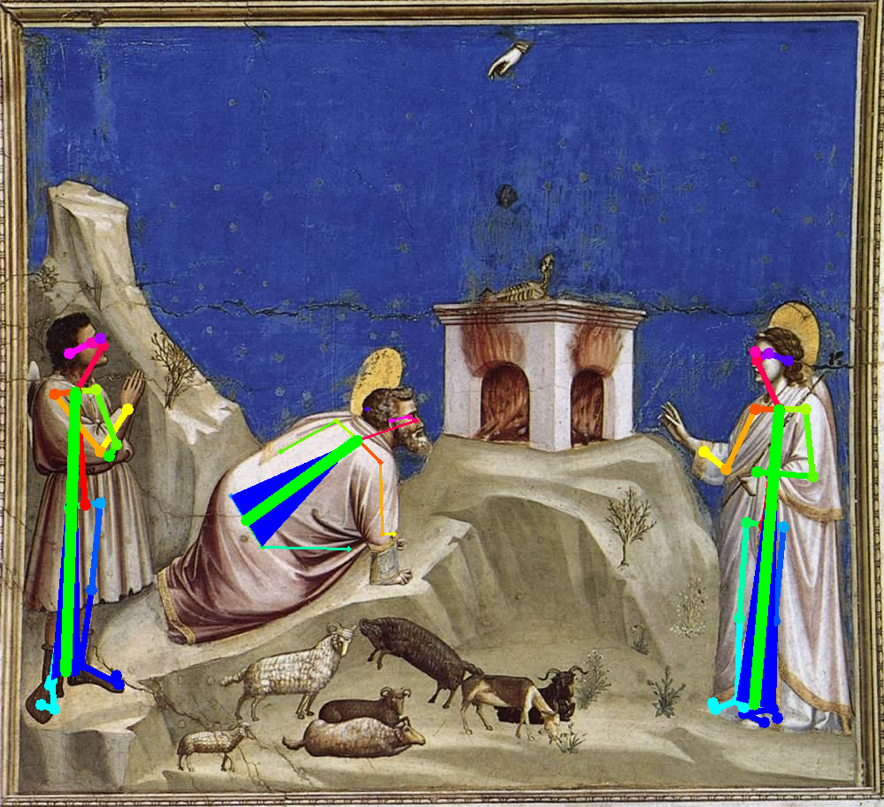}
		\caption{}
		\label{fig:pose-a}
	\end{subfigure}%
	\begin{subfigure}[t]{0.25\linewidth}
		\centering
		\includegraphics[width=0.9\linewidth, height=0.75\linewidth]{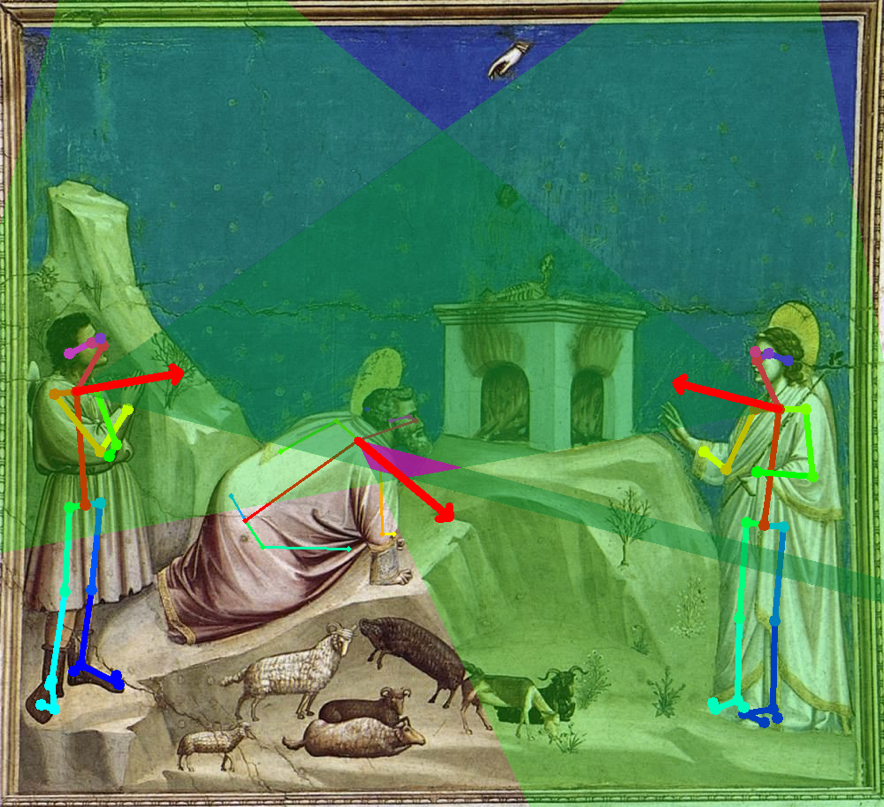}
		\caption{}
		\label{fig:pose-b}
	\end{subfigure}%
	\begin{subfigure}[t]{0.25\linewidth}
		\centering
		\includegraphics[width=0.9\linewidth, height=0.75\linewidth]{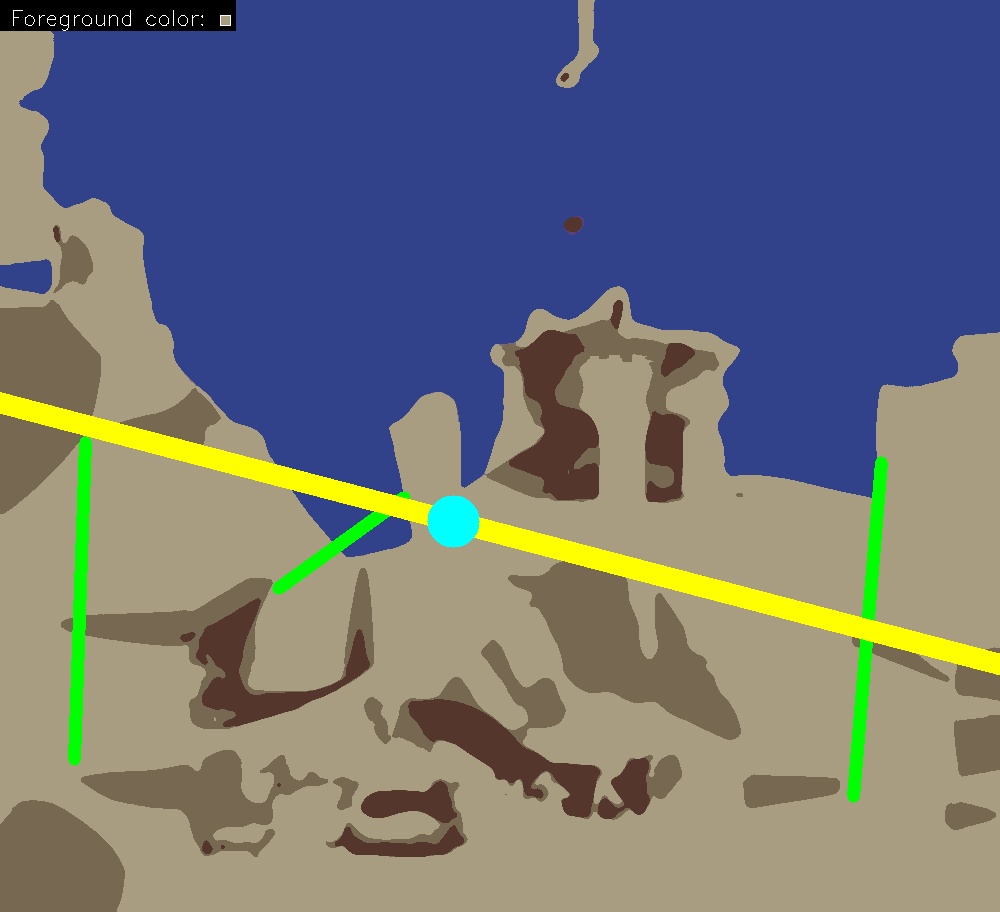}
		\caption{}
		\label{fig:pose-c}
	\end{subfigure}%
	\caption{\protect{\subref{fig:pose-d}} $25$ pose-keypoints used by OpenPose~\protect{\cite{caoOpenPoseRealtimeMultiPerson2019}}; \protect{\subref{fig:pose-e}} Pose triangle using groups of keypoints; \protect{\subref{fig:pose-a}} OpenPose generated keypoints, thicker lines have higher confidence. Generated \emph{Pose Triangles} (\protect{\cref{sec:PoseTriangleAbstraction}}a) in blue and the final abstracted result in bold green line; \protect{\subref{fig:pose-b}} Gaze-bisection-vectors (\protect{\cref{sec:gazebivec}}c) in bold red, cones in light transparent green, the intersection of all cones in purple; \protect{\subref{fig:pose-c}} Overlapping intersections in purple}
	\label{fig:pose}
\end{figure}

In order to detect ALs and PLs, we first pass the image through pre-trained OpenPose network. The output is a $25$-dimensional keypoint vector shown in \cref{fig:pose-d}. It is easy for OpenPose to detect all the keypoints for a fully visible human body. However, sometimes the full body is not visible, as can be seen for the character in the middle in \cref{fig:pose-a}. We correct such poses using a pose corrector (\cref{sec:PoseTriangleAbstraction}a). We then detect the gazes of these characters using three major keypoints detected by OpenPose (\cref{sec:GazeEstCorr}b). The output of the gaze estimator gives us the gaze directions in the form of cones as visualized in \cref{fig:pose-b}. The intersecting region of all the gaze cones is considered as the region of interest, where the main ARs would be present. Based on the intersection pattern of the cones, there can be multiple ARs. Also, a line representing the direction and intersection of all the views is considered to be the global AL. The slope of this line is calculated by combining all the gaze slopes (\cref{sec:gazebivec}c).
    
\subsubsection{(a) Pose Correction}\label{sec:PoseTriangleAbstraction}
For pose estimation, we tried a few SOTA methods, including associative embedding for joint detection and grouping~\cite{newell2017associative}, however OpenPose~\cite{caoOpenPoseRealtimeMultiPerson2019} gave the best results. Bottom-up approaches have the overhead of detecting persons in art historical images~\cite{artchars}, which is an entirely different problem altogether. Hence, we use OpenPose~\cite{caoOpenPoseRealtimeMultiPerson2019} pre-trained on the Human Foot
keypoint dataset~\cite{caoOpenPoseRealtimeMultiPerson2019} in combination with the 2017 version of
the COCO~\cite{linMicrosoftCOCOCommon2015} dataset. The network takes an entire image as input and returns fully connected body poses for all humans detected in an image as output.

Often OpenPose would predict incorrect keypoints, which were difficult to interpret, motivating us to come up with the pose-triangles approach which is robust for person interpretation irrespective of the occluded keypoints. With the detected keypoints, we create a pose triangle. For the triangle corners, we split the $25$ keypoints into three body regions as shown in \cref{fig:pose-e}. The region from shoulders towards the head forms the top triangle corner (brown trapezoid). The left (green trapezoid) and the right leg (yellow trapezoid) regions form the left and the right corners of the pose triangle, respectively. The specific keypoints associated with these regions are top$_C$ $\leftarrow$ [0,1,2,5,15,16,17,18], left$_C$ $\leftarrow$ [9,10,11,22,23,24], right$_C$ $\leftarrow$ [12,13,14,19,20,21].
The keypoint from each region with the highest confidence score is chosen to be the representative point from which the pose triangle will be formed. For each of the pose triangles, we create a bisector line called \emph{pose line} (PL) from the top corner of the triangle to the line segment formed by the other two corners (bold green lines as seen in \cref{fig:pose-a}). The keypoints' splits ($3$ regions for $3$ corners of the pose triangle) are chosen heuristically.

\subsubsection{(b) Gaze Estimation + Correction}\label{sec:GazeEstCorr}
Our data did not have annotations in order to fine-tune an existing gaze estimation model. Also, our goal was to get a rough estimate of the gazes for all the persons in order to detect the central focus within the image. Looking at the pose keypoints, we can argue that the gaze direction can roughly be estimated from the face and neck keypoints. Here, we exploit this hypothesis and use the pose keypoints $0, 1$ and $8$ to generate a bisection vector as the first step (red vector, \cref{fig:pose-b}). The bisection vector bisects the line segment joining keypoints $0$ and $8$. In \cref{fig:pose-b}, we observe that these estimates are few degrees off of the original gaze and hence we also apply a correction to the gaze vector, denoted as correction angle. We skip those cases when one of the three keypoints is missing. Rather than just viewing the exact gaze, we represent gazes using gaze cones with an opening angle of 50 (25 degrees +/- direction) degrees in order to compensate the error estimate (green transparent cones, \cref{fig:pose-b}). Experiments with various angles however did not affect the ARs and hence, we heuristically chose to keep the value 50\footnote{similar results were achieved by using various angles: 10, 20, 60, 80.}.
    
The intersection of these cones are the probable regions where everyone in the scene is looking at. Hence, the centroids of these intersections become the starting points of our global ALs. The cases with no intersection are skipped. In cases where there is more than one intersecting area, we proceed with two or more ARs (\cf \cref{fig:circlemap-b,fig:circlemap-c}). In order to plot the AL on our canvas, we require its slope. Therefore, we aggregate all the individual gaze directions in order to find the slope of the global AL. 

\subsubsection{(c) Combination of all gaze-bisection-vector slopes}\label{sec:gazebivec}
\begin{figure}[t]
		\centering
		\begin{subfigure}[t]{0.25\linewidth}
			\centering
			\includegraphics[width=0.9\linewidth]{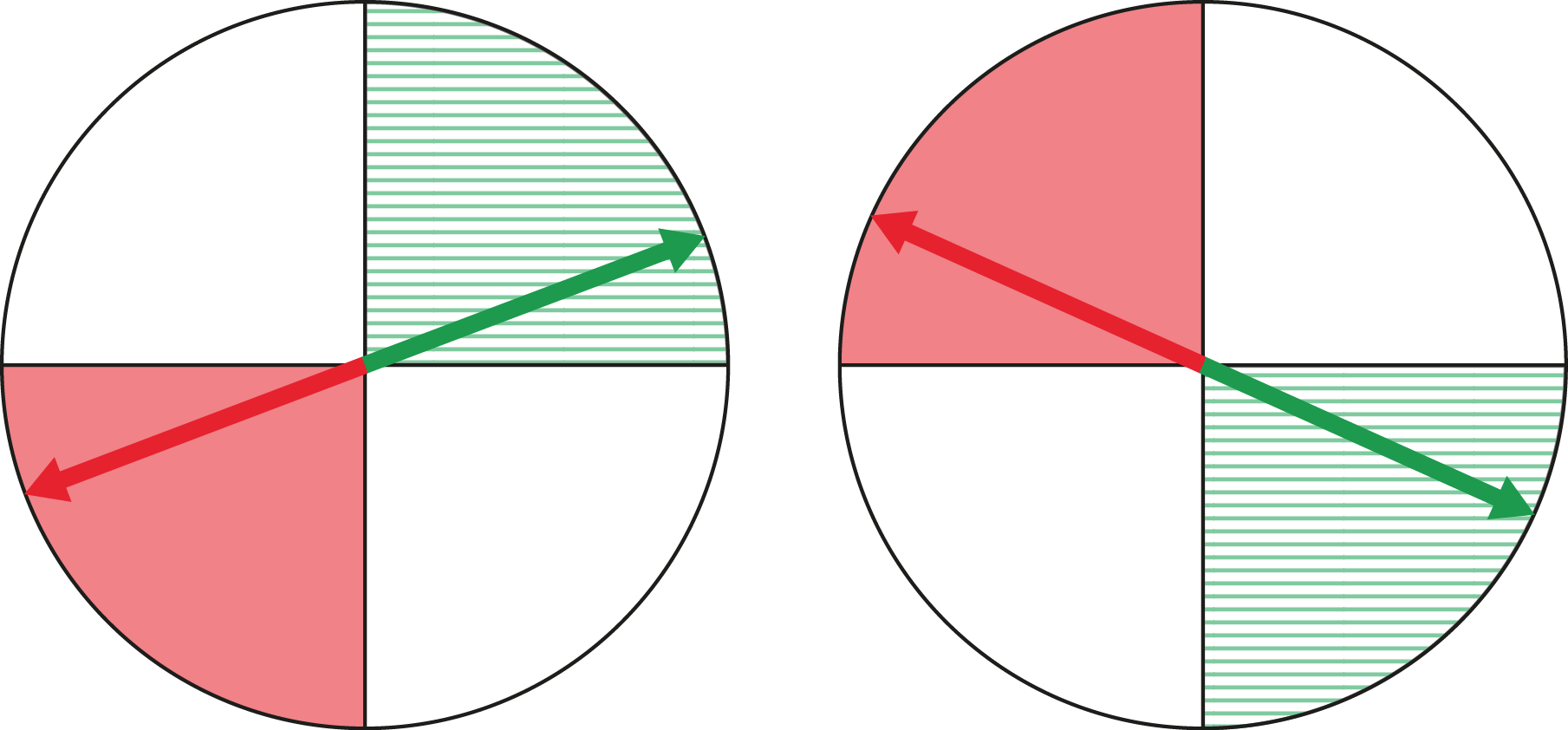}
			\caption{}
			\label{fig:circlemap-a}
		\end{subfigure}%
		\begin{subfigure}[t]{0.25\linewidth}
			\centering
			\includegraphics[width=0.85\linewidth, height=0.75\linewidth]{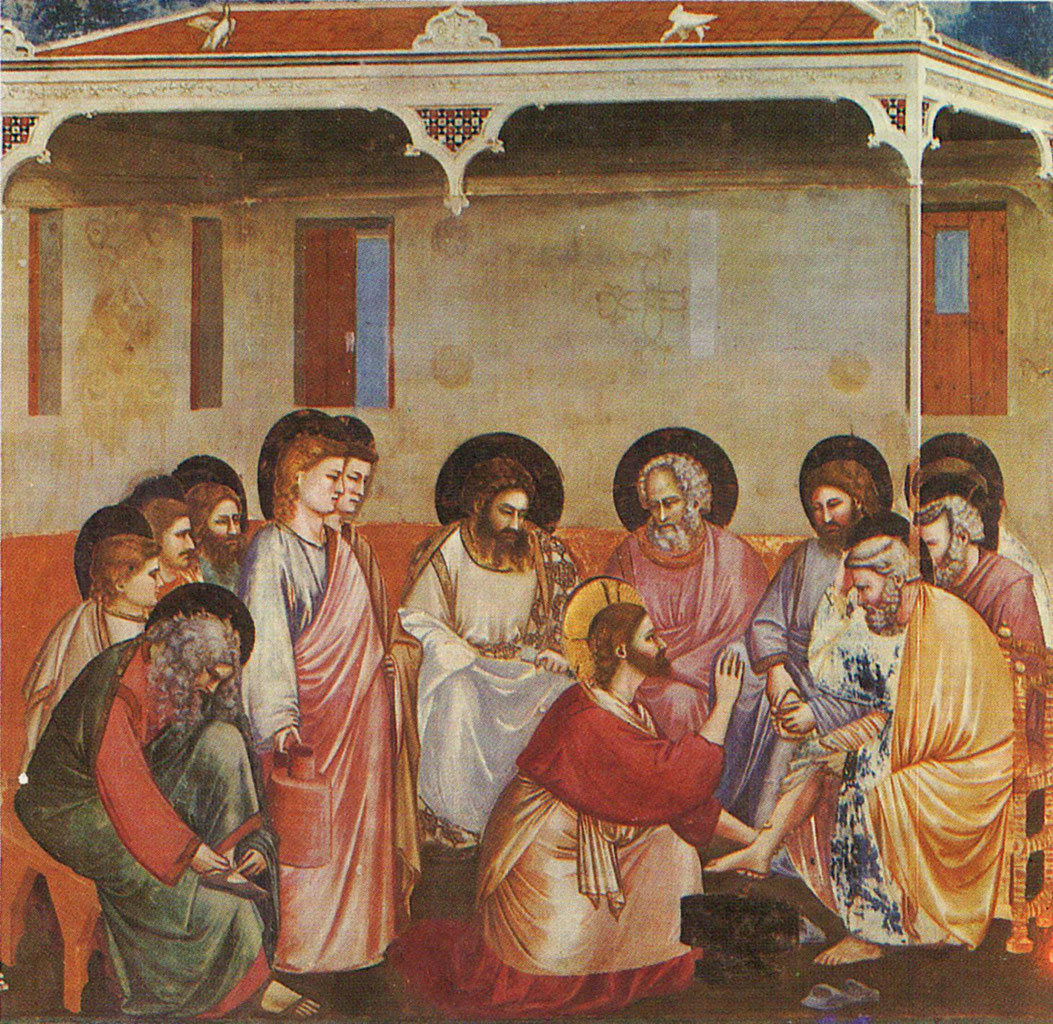}
			\caption{}
			\label{fig:circlemap-b}
		\end{subfigure}%
		\begin{subfigure}[t]{0.25\linewidth}
			\centering
			\includegraphics[width=0.85\linewidth, height=0.75\linewidth]{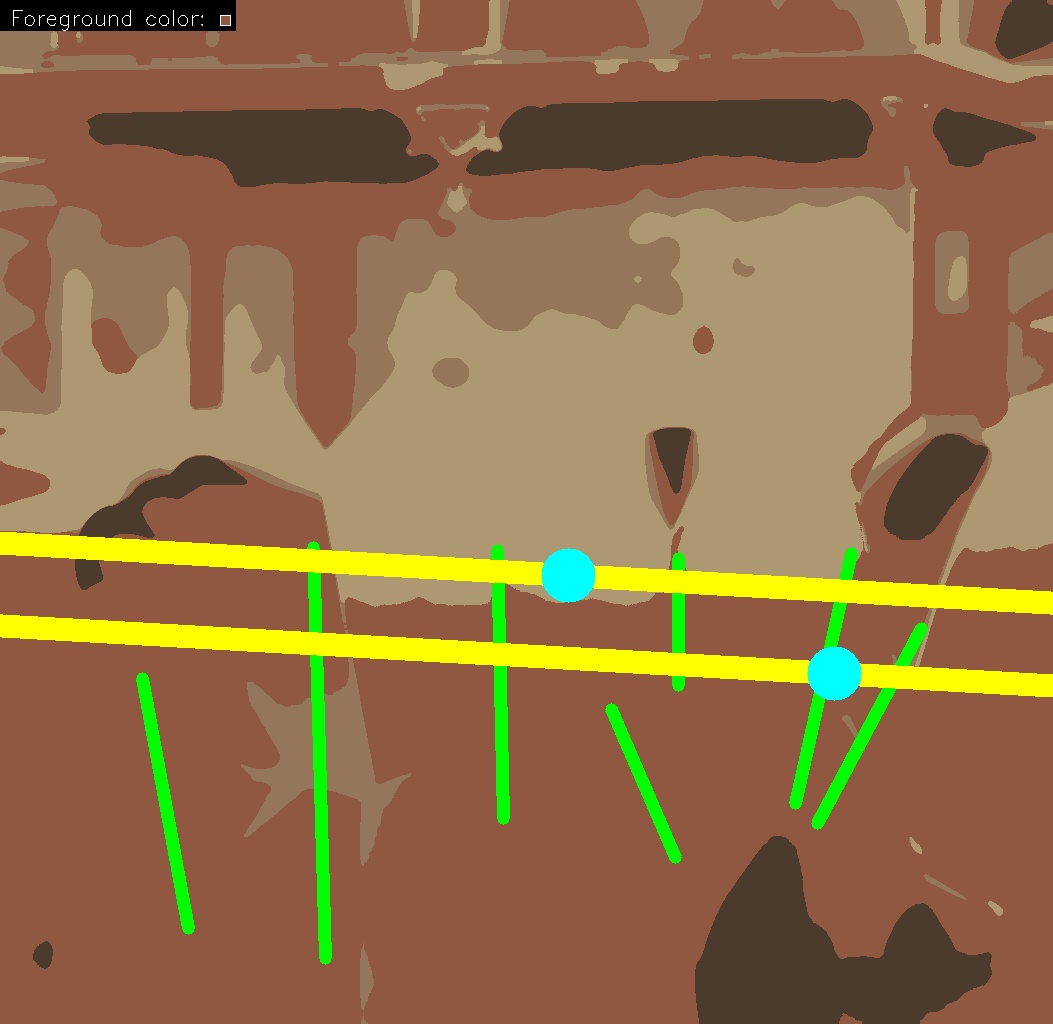}
			\caption{}
			\label{fig:circlemap-c}
		\end{subfigure}%
		\begin{subfigure}[t]{0.25\linewidth}
			\centering
			\includegraphics[width=0.85\linewidth, height=0.75\linewidth]{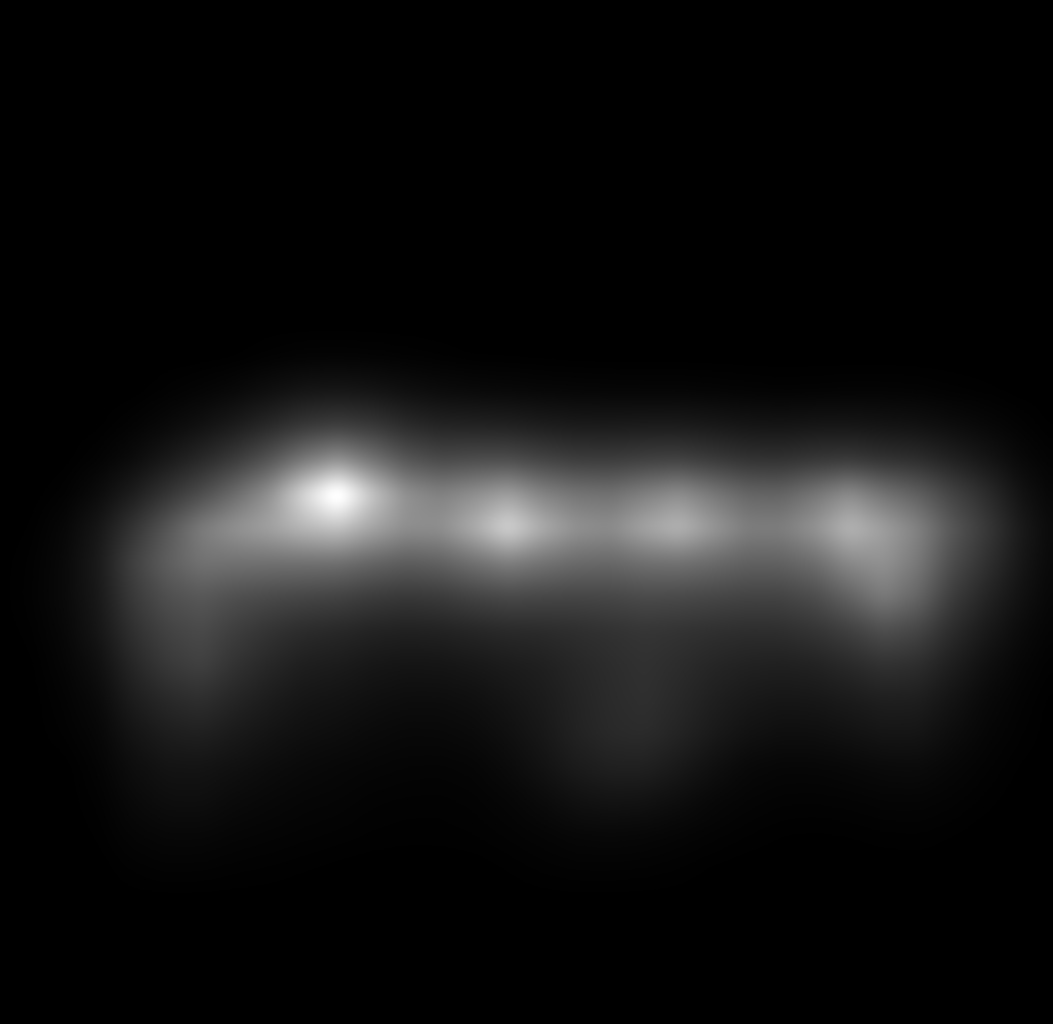}
			\caption{}
			\label{fig:salientfuss}
		\end{subfigure}%
\caption{\protect{\subref{fig:circlemap-a}} Gaze-Bisection-Vectors (red vectors) that point towards \nth{2} and \nth{3} quadrants are mapped to \nth{4} and \nth{1} quadrants (green vectors), respectively. \protect{\subref{fig:circlemap-b}}, \protect{\subref{fig:circlemap-c}} show an example where there are $2$ global action lines and $2$ action regions detected. \protect{\subref{fig:salientfuss}} is the eye-fixation map~\protect{\cite{cornia2018predicting}} of the image.}
        \label{fig:circlemap}
	\end{figure}

For calculating the slope of our global ALs, we aggregate all individual gaze directions (average of all the individual gaze slope) and find a common slope, where only the slopes of the gaze-bisection-vectors are considered. The slope value is with respect to the x-axis (Cartesian Coordinates) in the positive horizontal direction. Since the slope remains the same when we rotate the gaze-bisection-vector by $180$ degrees, we map all the gaze vectors pointing towards the \nth{2} and \nth{3} quadrant to \nth{4} and \nth{1} quadrants respectively (\cref{fig:circlemap-a}). We then draw a line in our output canvas through every centroid of the previously calculated intersection areas and use this aggregated slope as the slope of this new line. \cref{fig:pose-b,fig:pose-c} show how the aggregated gaze vectors form the global AL.  

\subsection{Semantic Foreground/Background separation}\label{sec:backfore}
While analyzing human behavior within the semantics of scene understanding, humans present in the scene usually form the foreground. We exploit this property of scene semantics in our approach and consider the humans as our foreground (FG). Additionally, we define the objects near the human poses and the immediate surroundings as part of their foreground and the rest of the image as background (BG). The pipeline for FG and BG separation is shown in \cref{fig:pipeline}.
It consists of the following three steps, which are explained in more details below: image filtering, keypoint-based inpainting, and pose- and color-informed k-means clustering.  
\begin{figure}[t]
	\centering
	\begin{subfigure}[t]{0.24\linewidth}
		\centering
		\includegraphics[width=0.9\linewidth, height=0.85\linewidth]{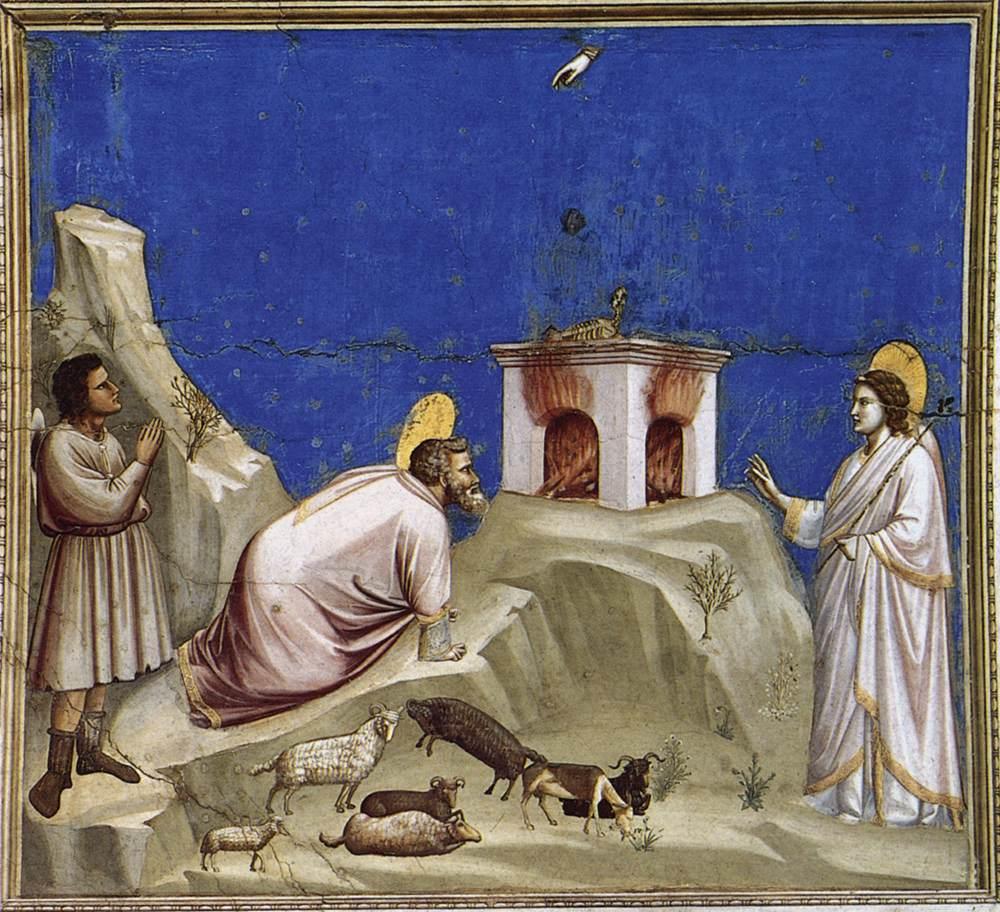}
		\caption{}
		\label{fig:fgbg-a}
	\end{subfigure}%
	\begin{subfigure}[t]{0.24\linewidth}
		\centering
		\includegraphics[width=0.9\linewidth, height=0.85\linewidth]{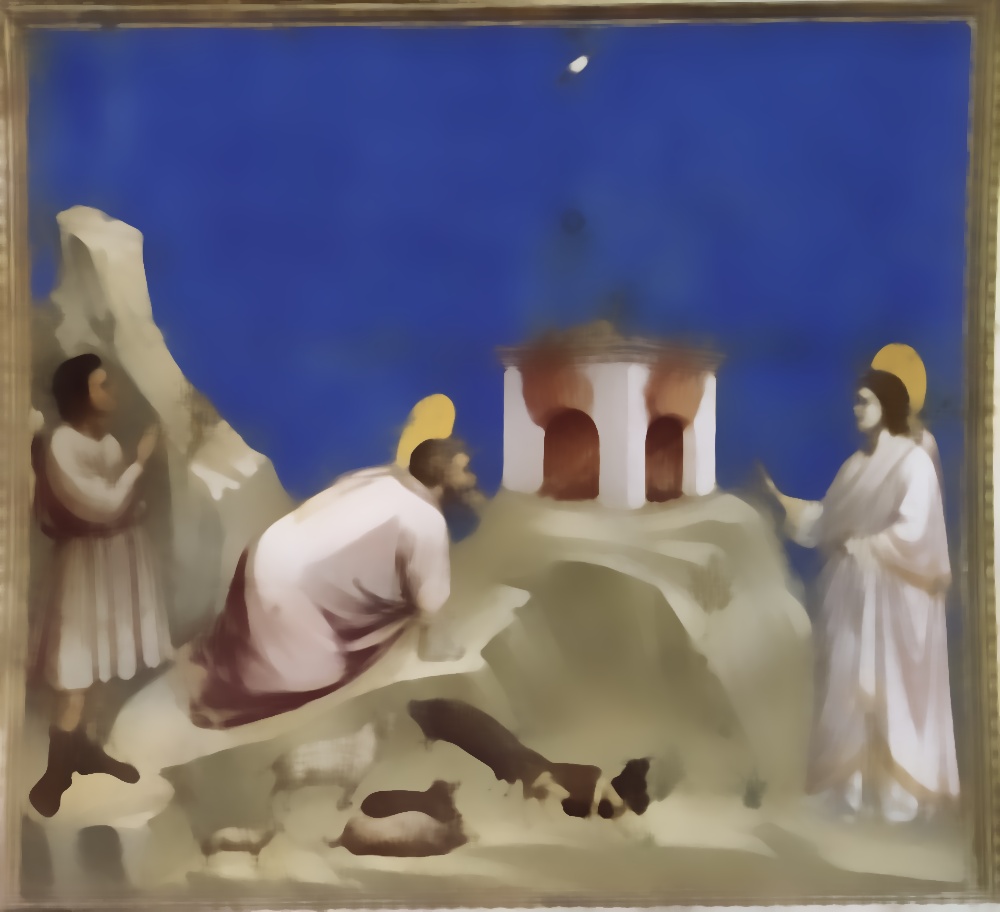}
		\caption{}
		\label{fig:fgbg-b}
	\end{subfigure}%
	\begin{subfigure}[t]{0.24\linewidth}
		\centering
		\includegraphics[width=0.9\linewidth, height=0.85\linewidth]{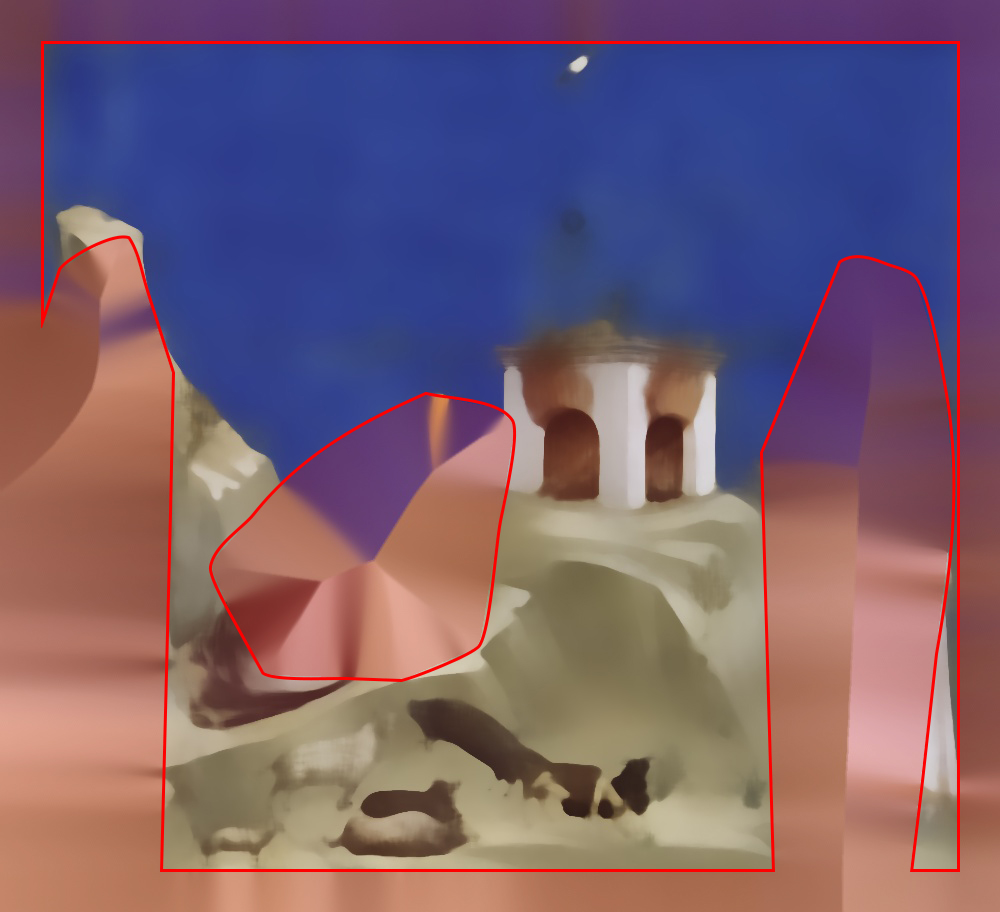}
		\caption{}
		\label{fig:fgbg-c}
	\end{subfigure}%
    \begin{subfigure}[t]{0.24\linewidth}
    	\centering
    	\includegraphics[width=0.9\linewidth, height=0.85\linewidth]{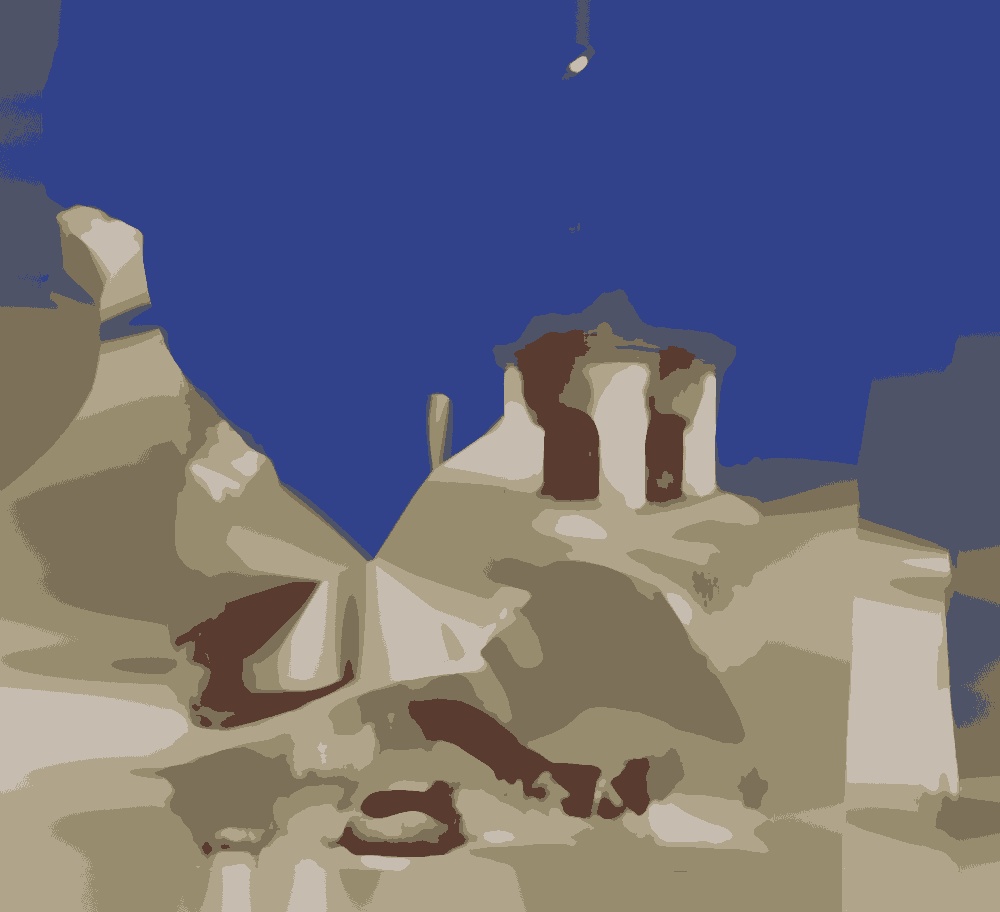}
    	\caption{}
    	\label{fig:fgbg-d}
    \end{subfigure} 
    
    \begin{subfigure}[t]{0.24\linewidth}
    	\centering
    	\includegraphics[width=0.9\linewidth, height=0.85\linewidth]{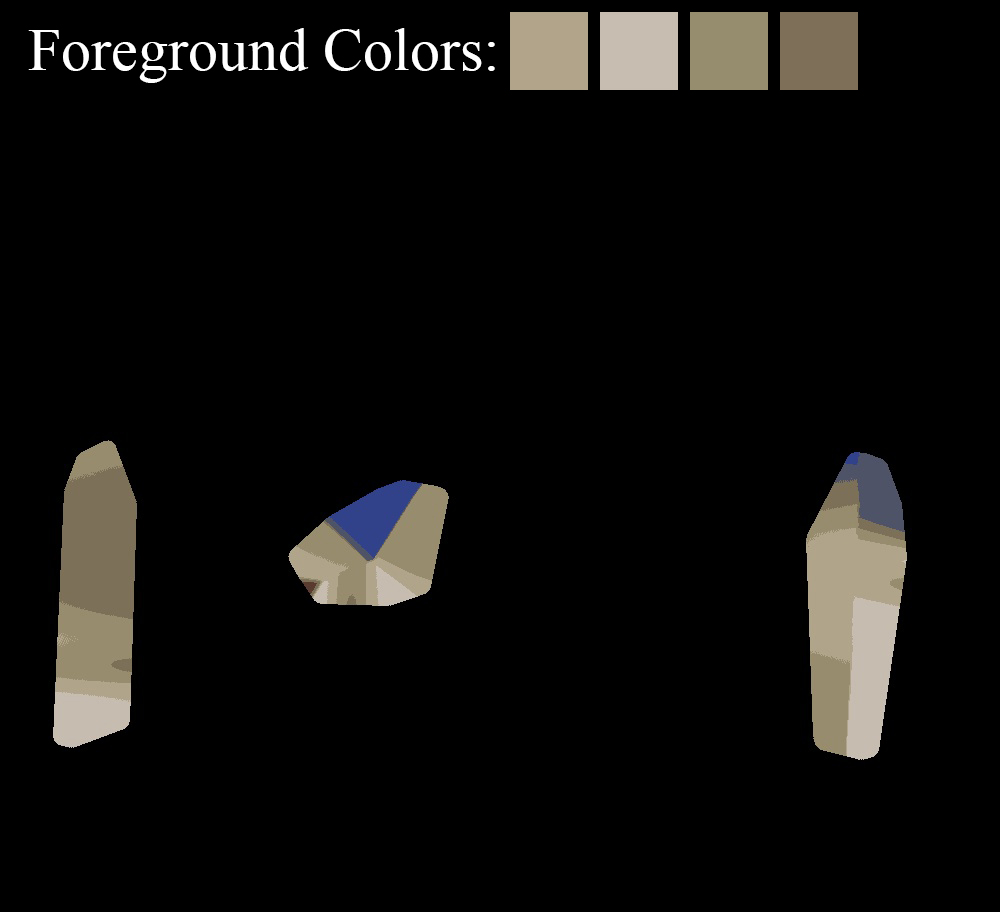}
    	\caption{}
    	\label{fig:fgbg-e}
    \end{subfigure}%
    \begin{subfigure}[t]{0.24\linewidth}
    	\centering
    	\includegraphics[width=0.9\linewidth, height=0.85\linewidth]{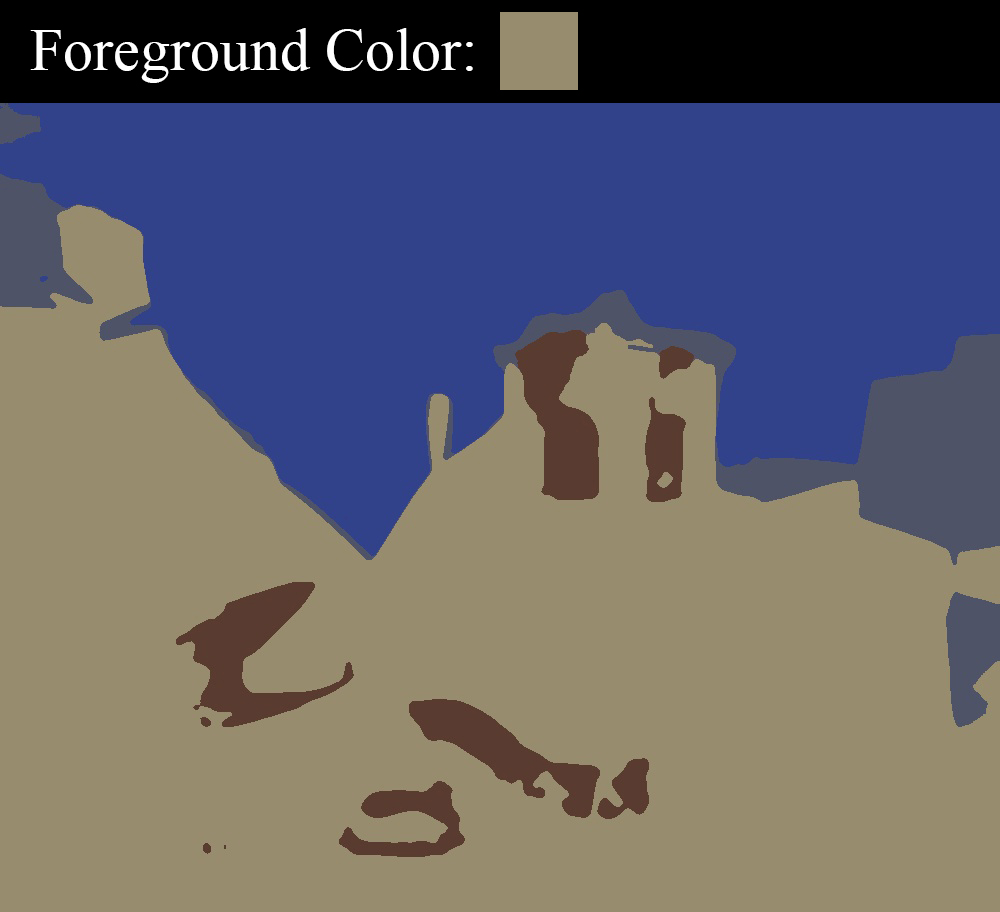}
    	\caption{}
    	\label{fig:fgbg-f}
    \end{subfigure}%
    \begin{subfigure}[t]{0.24\linewidth}
    	\centering
    	\includegraphics[width=0.9\linewidth, height=0.85\linewidth]{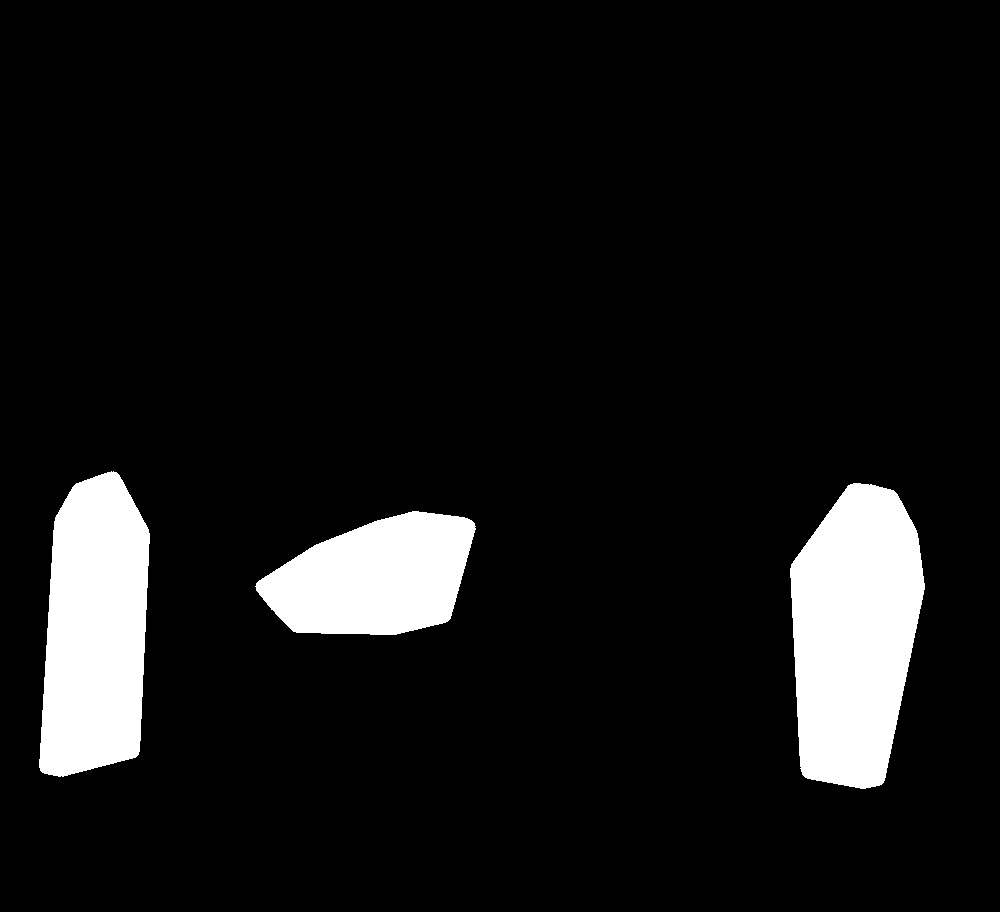}
    	\caption{}
    	\label{fig:fgbg-g}
    \end{subfigure}%
    \begin{subfigure}[t]{0.24\linewidth}
    	\centering
    	\includegraphics[width=0.9\linewidth, height=0.85\linewidth]{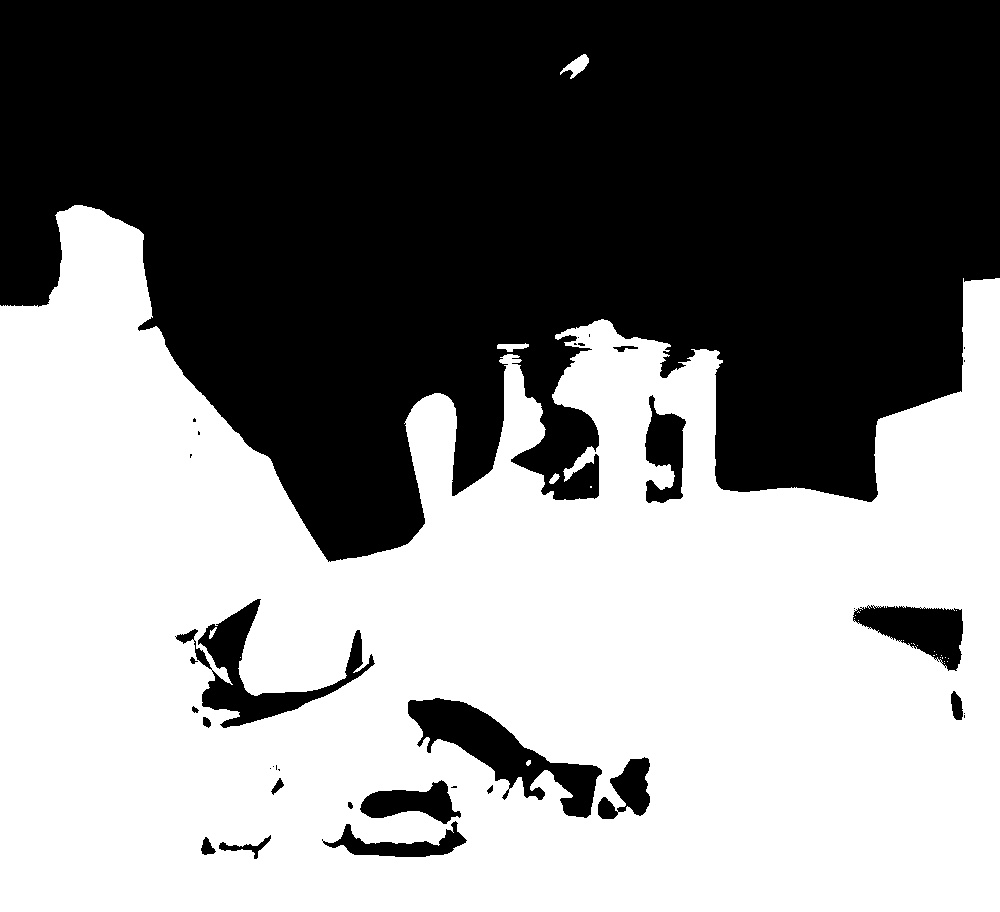}
    	\caption{}
    	\label{fig:fgbg-h}
    \end{subfigure}%
         \caption{Visual steps of \emph{ICC}: \protect{\subref{fig:fgbg-a}} Original image, \protect{\subref{fig:fgbg-b}} Cracks filtered, \protect{\subref{fig:fgbg-c}} Bodies and picture frame inpainted, \protect{\subref{fig:fgbg-d}} Colors clustered using k-means, \protect{\subref{fig:fgbg-e}} Dominant colors under body positions selected as foreground colors, blue colors are below the threshold, \protect{\subref{fig:fgbg-f}} Detected foreground colors replaced and output filters applied, \protect{\subref{fig:fgbg-g}} Binarization applied to \protect{\subref{fig:fgbg-e}}, \protect{\subref{fig:fgbg-h}} Binarized foreground/background separation.}
         \label{fig:fgbg}
    \end{figure}
	
\subsubsection{(a) Image filtering.}\label{sec:imfilt}
We observe that historical paintings show chipped paint, cracks or weathering, so we first apply a median filter that helps in reducing these artifacts. We apply a bilateral filter to preserve the sharp edges as they are most crucial to separate major objects and protagonists as seen in \cref{fig:fgbg-b}. 
\subsubsection{(b) Keypoint-based Inpainting.}\label{sec:inpaint}
Next, we inpaint all pixels covered by detected human poses. Therefore, we generate the convex hull of all pose keypoints and scale the hull up by \SI{70}{\percent} in x-direction and \SI{40}{\percent} in y-direction, to ensure the mask is covering a little more than the human body in the painting. Using this mask, we inpaint the previously filtered image using the fast marching method~\cite{teleaImageInpaintingTechnique2004}. \Cref{fig:fgbg-c} shows human bodies being replaced with a color mix marked by the red mask. These colors are considered as the potential foreground colors. 
\subsubsection{(c) Pose- and Color-informed k-means clustering.}\label{sec:kmeans}
The inpainted image is then clustered into a smaller amount of colors using k-means clustering. This creates smaller color clusters (\cf~\cref{fig:fgbg-d}). The colors near the body postures of the characters are associated with the foreground by virtue of being related to the bodies of the characters. Next, we generate a second mask around the human poses, but scale each hull down by \SI{30}{\percent} in y-direction to ensure that our mask is only placed over the core of our previously generated color mix (\cf~\cref{fig:fgbg-c}) using the inpainting (\cf~\cref{fig:fgbg-e}). All colors under this mask covering more than \SI{6}{\percent} of the mask are then considered as foreground color. For experiments, we tried k=3, 5, 7, 9 and found k=7 to generate better qualitative results. The scaling factors for the convex hull were chosen heuristically. 

\subsection{Generating Output Canvas -- Bringing It All Together}\label{sec:icc}
We generate two types of ICC, one colored ICC and a binary one denoting foreground/background pixels. In the colored version, all detected foreground colors are replaced with the most dominant foreground color. We apply median filter to remove small leftover color blobs/fragments as a post-processing step. For the binary version, each pixel is set to one if it is one of the foreground colors and zero otherwise. We apply morphological dilation and erosion filtering on the binary version to perform the closing of small fragments as post-processing. Finally, a morphological filter opening is applied to remove small blobs in the background that are typically due to the k-means results on the clustered cracks. We then gather the ALs, ARs (\cref{sec:actions}) and overlay it on the canvas with FG/BG separation (\cref{sec:backfore}) to obtain our ICC. We can observe colored canvas in the \cref{fig:main-g}; and the binarized canvas in \cref{fig:main-h}.
    
\section{Evaluation and Experiments}\label{Disc}
In this section, we first discuss the user study done for quantitative evaluation and the metrics being used for the same. Then we do a qualitative interpretation of our method, followed by general validity of our method by testing on cross-domain data. In the last part, we give performance evaluation of our method based on the user study. 

\subsection{User Study -- Design and Evaluation Metrics}\label{sec:userstudy}
\subsubsection{Design} We evaluate the performance of ICC quantitatively and qualitatively through a user study. A set of 11 images were collected, 6 are Giotto's paintings, an annunciation scene, a baptism scene and 3 are chosen from COCO~\cite{linMicrosoftCOCOCommon2015} with people present in them. We asked the annotators to label global action lines (AL), action regions (AR) and pose lines (PL) on these images. For quality control, a demo of the labeling process with an example and associated instructions were continuously available to the annotators while doing the task. This video~\href{https://streamable.com/wk8ol8}{(https://streamable.com/wk8ol8)} describes how the user-study was conducted, including the labeling interface, the definition of each label and instructions to generate them with a sample example. We have 2 set of annotators, domain experts (E) and non-experts (NE). An $E$ has a background in art history  or its methodology; \emph{NE}s are all university graduates. A total of $72$ annotators ($10$~Es and $62$~NEs) were presented with this set of images and asked to label the AL, AR and PL in them. 
\begin{figure}[t]
	\centering
	\begin{subfigure}[t]{0.2\linewidth}
		\centering
		\includegraphics[width=0.95\linewidth,keepaspectratio]{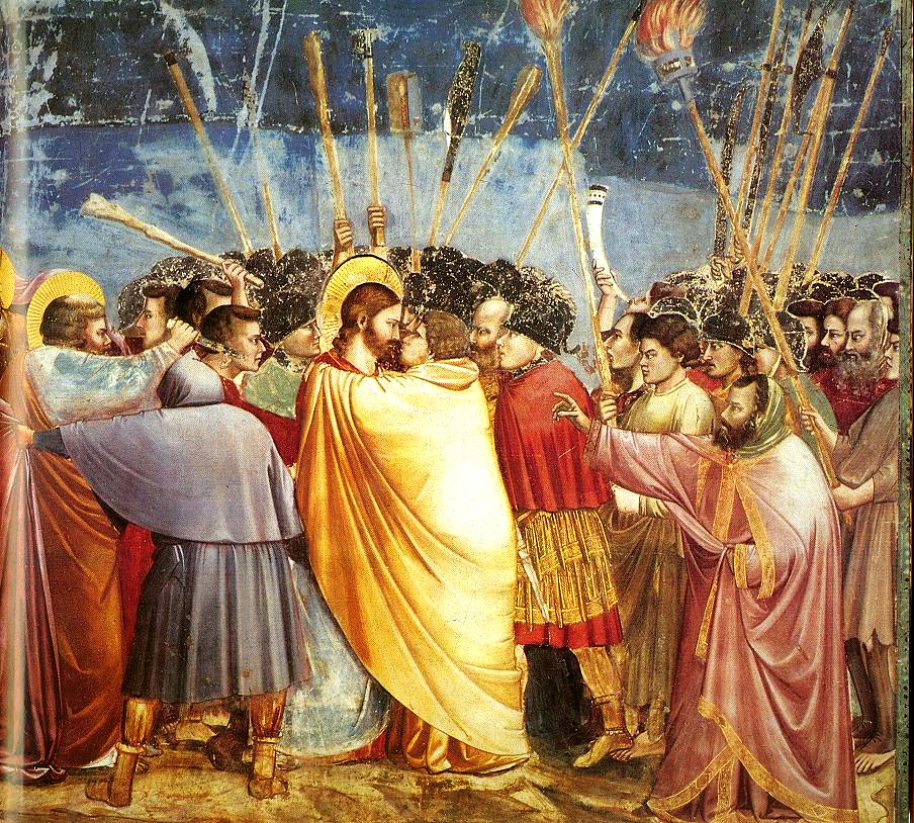}
		\label{fig:userstudy-a}
	\end{subfigure}%
	\begin{subfigure}[t]{0.2\linewidth}
		\centering
		\includegraphics[width=0.95\linewidth,keepaspectratio]{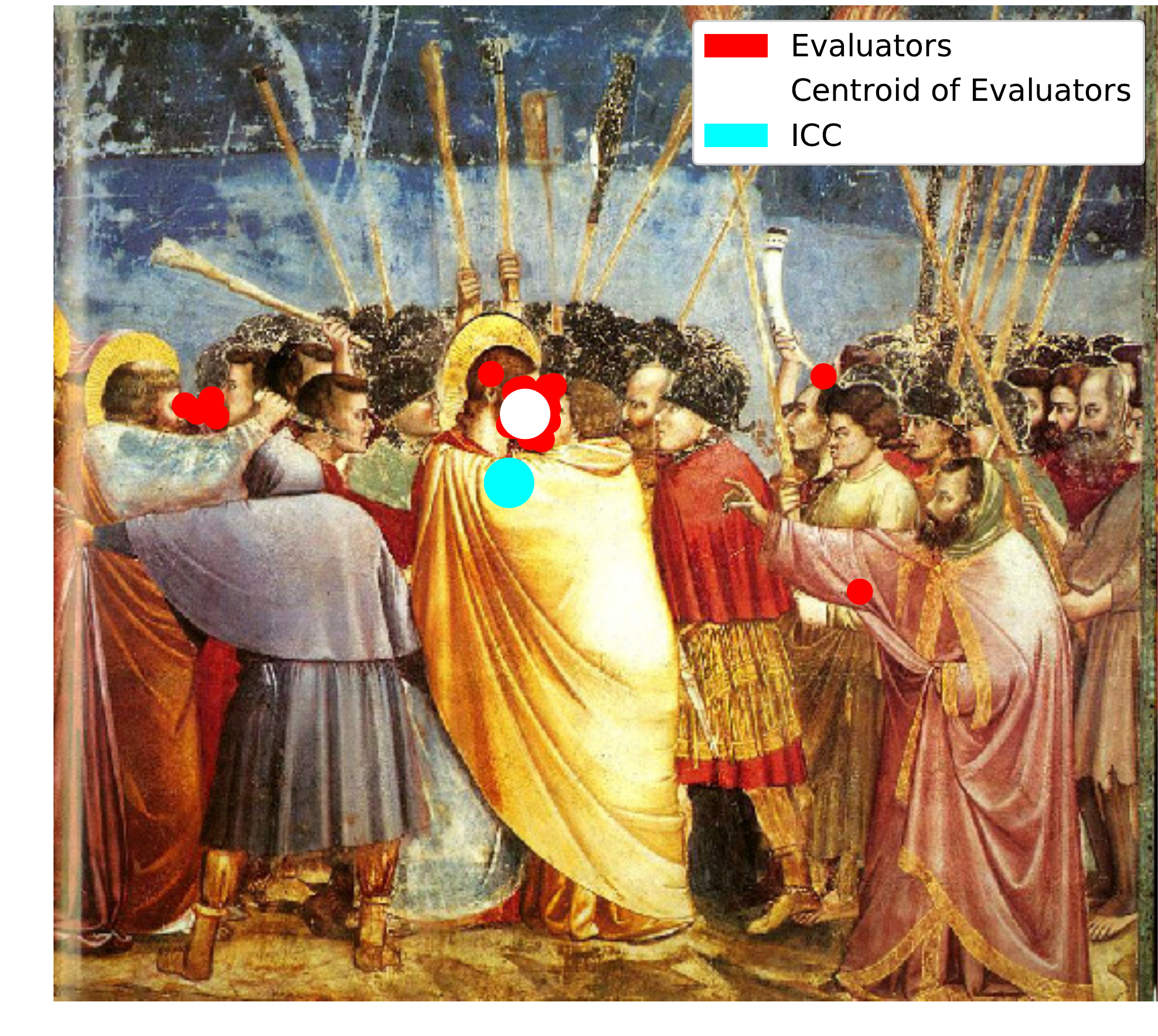}
		\label{fig:userstudy-b}
	\end{subfigure}%
	\begin{subfigure}[t]{0.2\linewidth}
		\centering
		\includegraphics[width=0.95\linewidth,keepaspectratio]{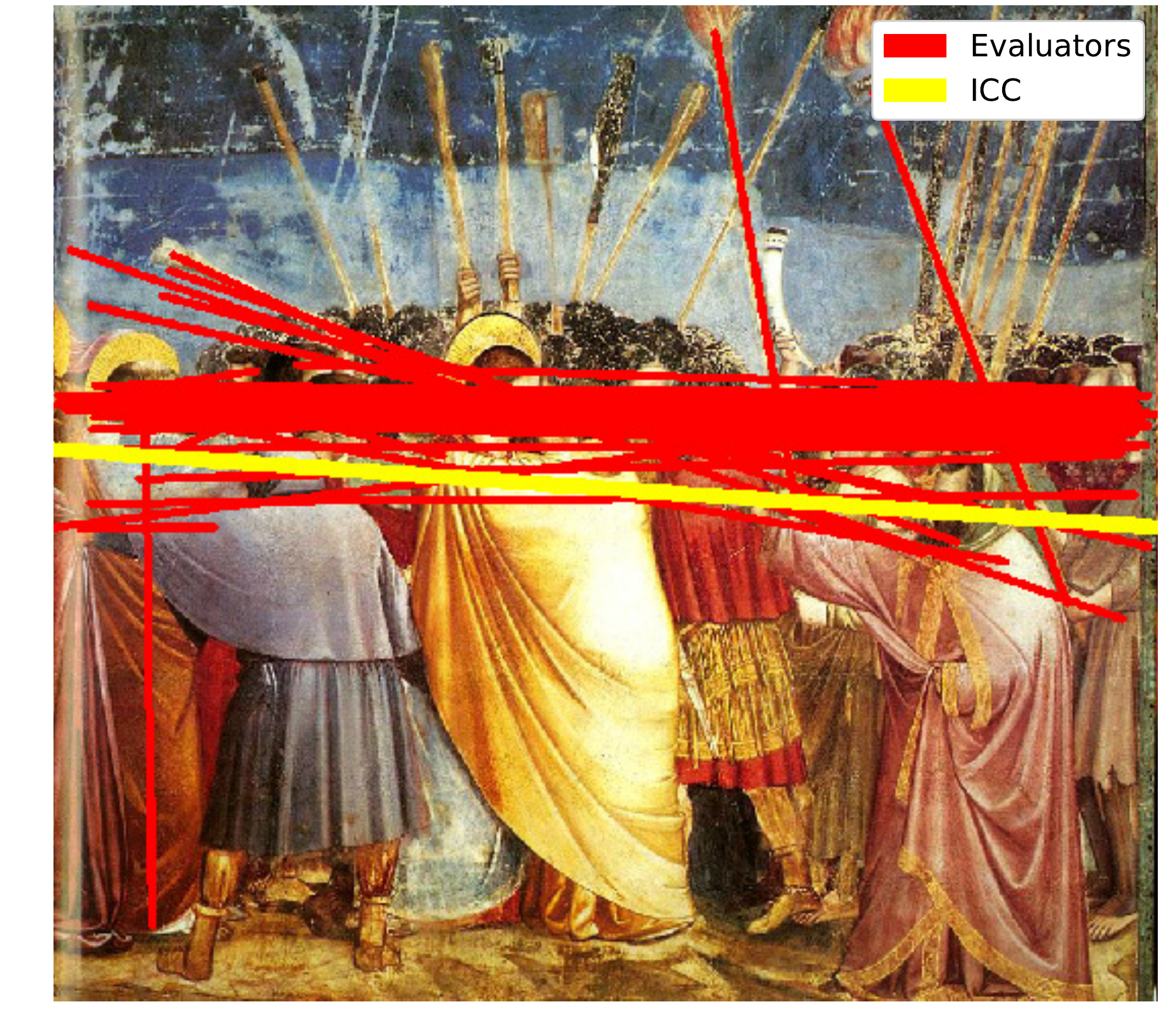}
		\label{fig:userstudy-c}
	\end{subfigure}%
	\begin{subfigure}[t]{0.2\linewidth}
		\centering
		\includegraphics[width=0.95\linewidth,keepaspectratio]{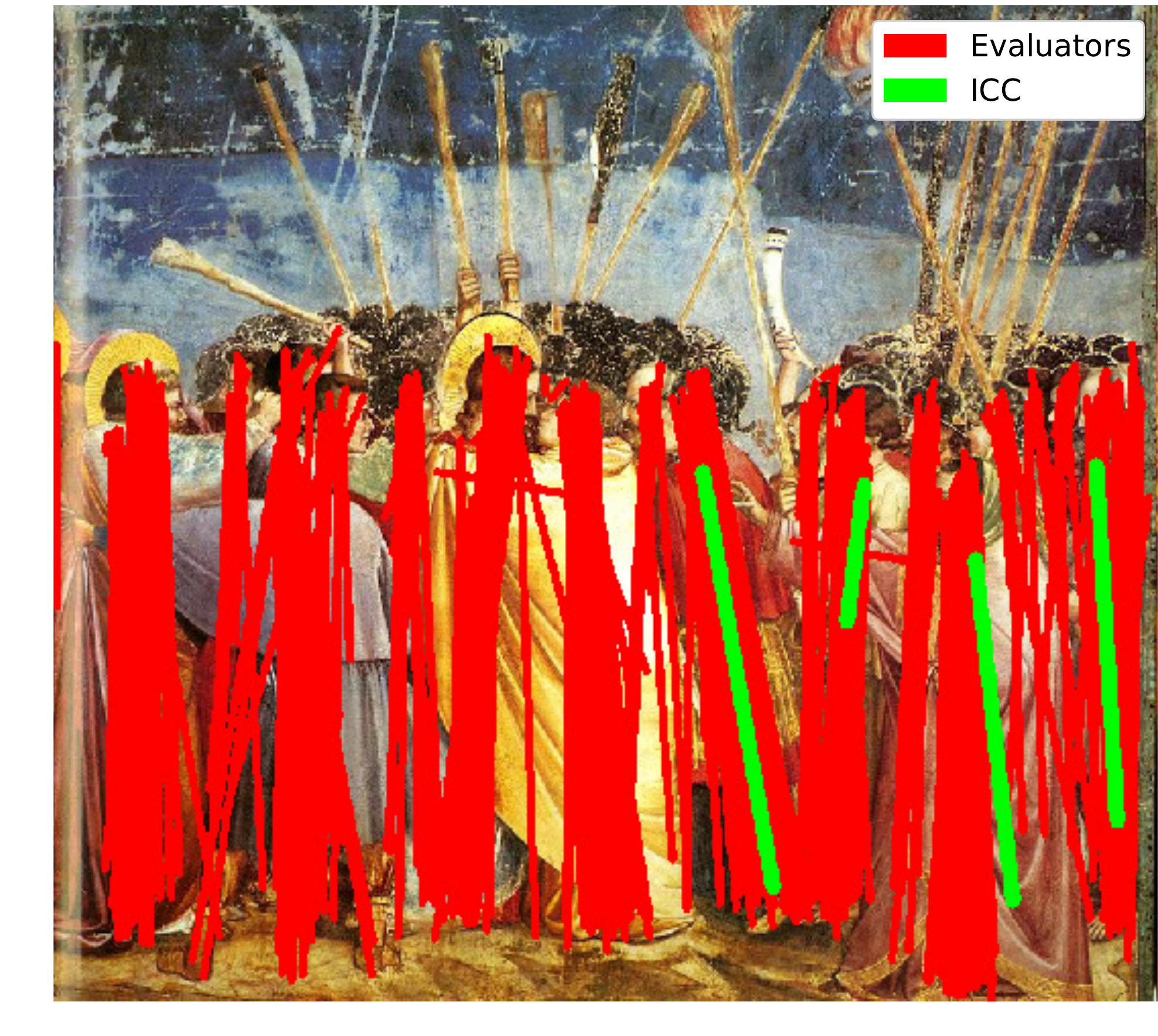}
		\label{fig:userstudy-d}
	\end{subfigure}%
	\begin{subfigure}[t]{0.2\linewidth}
		\centering
		\includegraphics[width=0.95\linewidth,keepaspectratio]{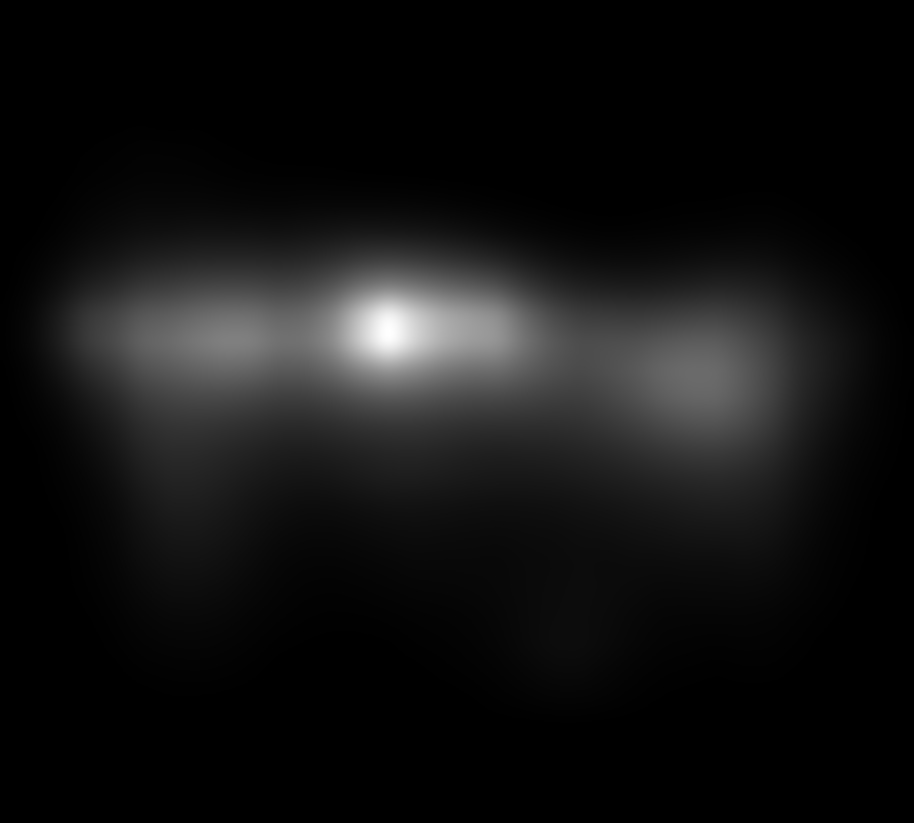}
		\label{fig:salientbap}
	\end{subfigure}	
	\begin{subfigure}[t]{0.2\linewidth}
		\centering
		\includegraphics[width=0.95\linewidth,keepaspectratio]{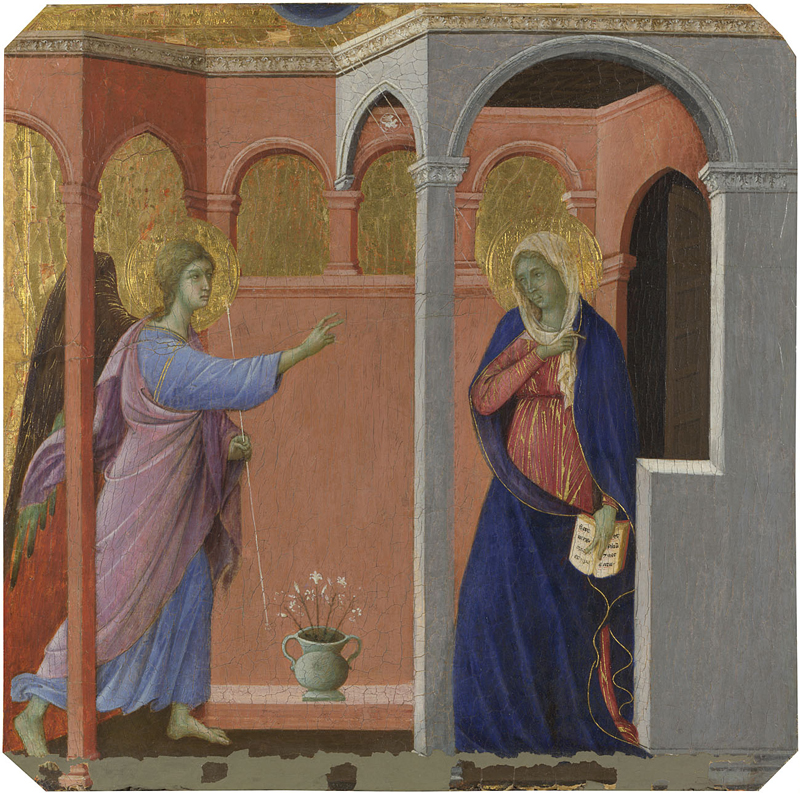}
		\label{fig:userstudy-e}
	\end{subfigure}%
	\begin{subfigure}[t]{0.2\linewidth}
		\centering
		\includegraphics[width=0.95\linewidth,keepaspectratio]{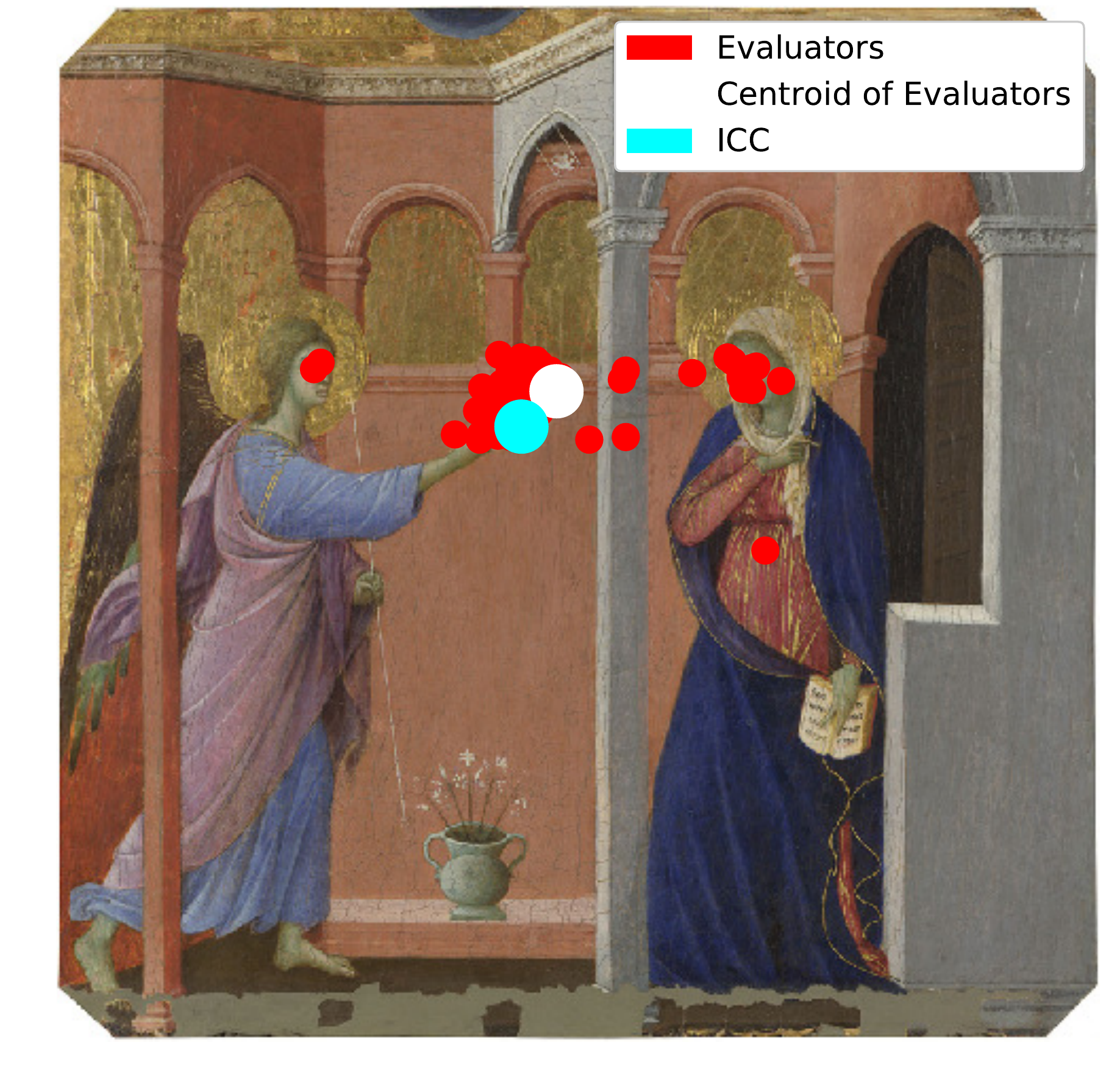}
		\label{fig:userstudy-f}
	\end{subfigure}%
	\begin{subfigure}[t]{0.2\linewidth}
		\centering
		\includegraphics[width=0.95\linewidth,keepaspectratio]{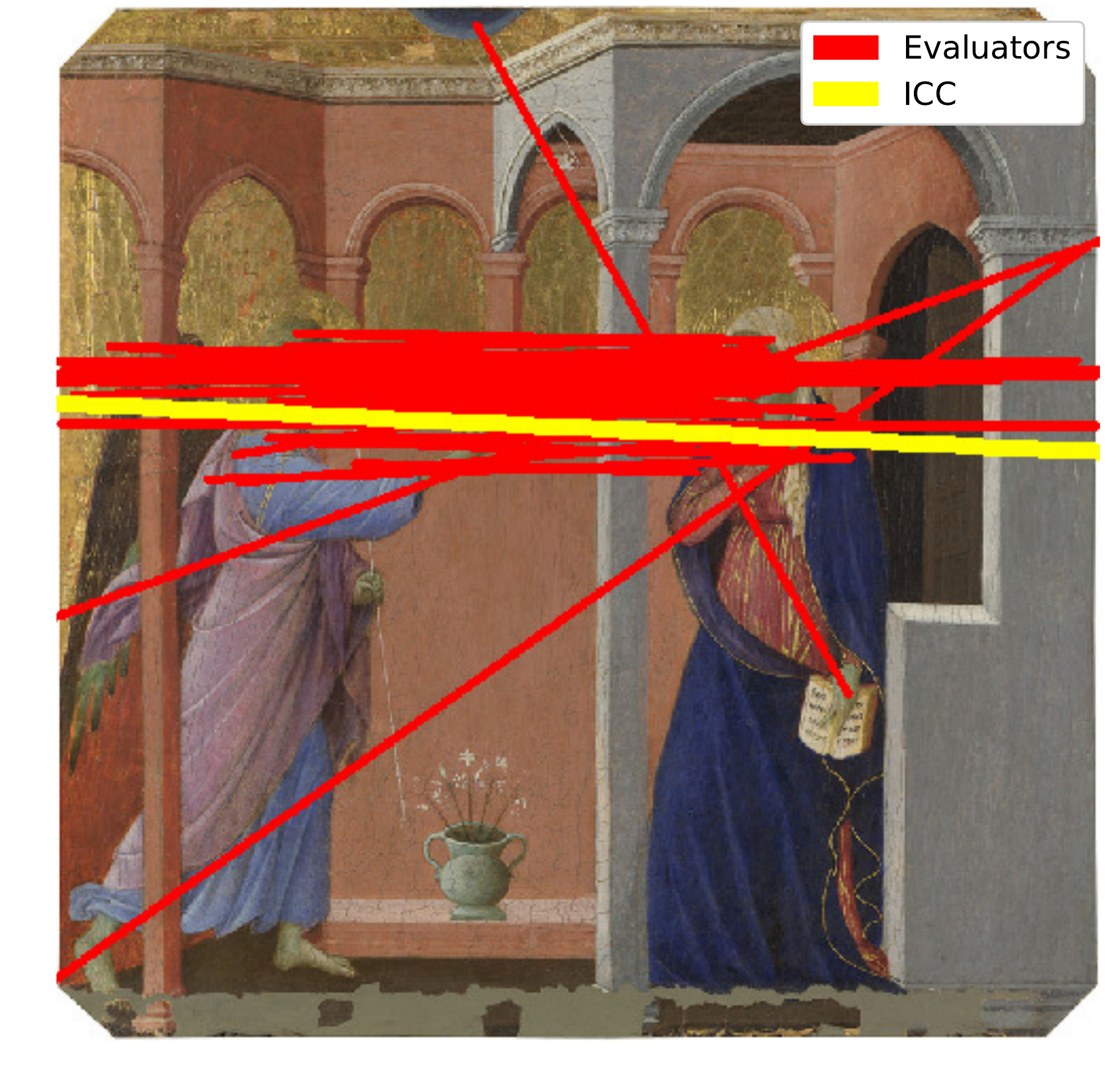}
		\label{fig:userstudy-g}
	\end{subfigure}%
	\begin{subfigure}[t]{0.2\linewidth}
		\centering
		\includegraphics[width=0.95\linewidth,keepaspectratio]{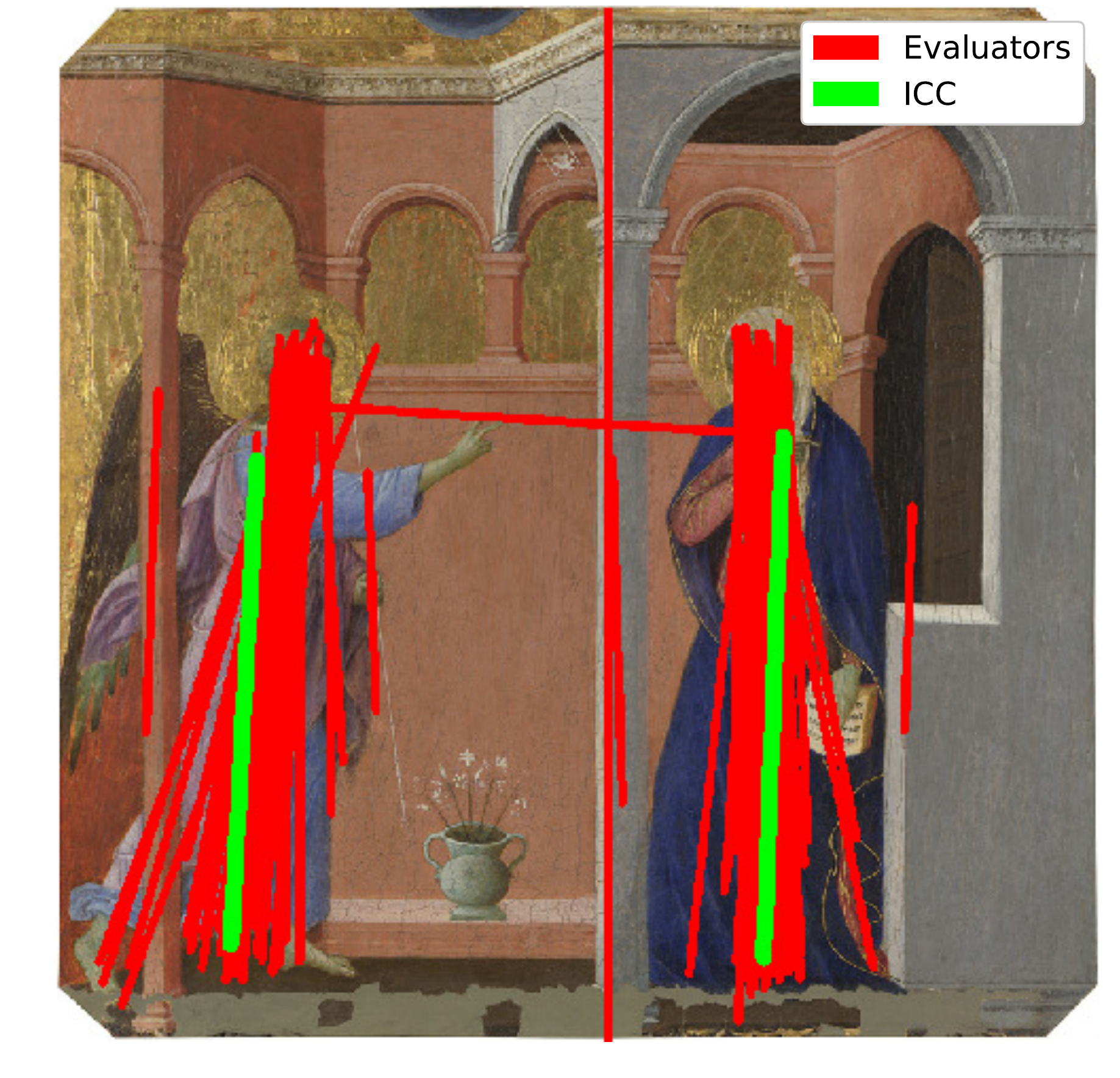}
		\label{fig:userstudy-h}
	\end{subfigure}%
	\begin{subfigure}[t]{0.2\linewidth}
		\centering
		\includegraphics[width=0.95\linewidth,keepaspectratio]{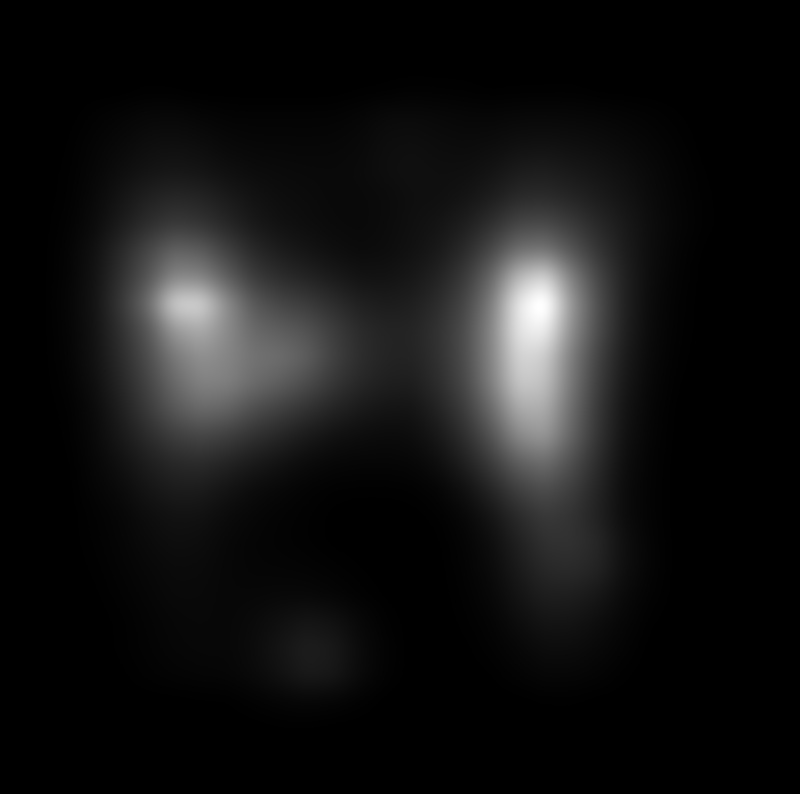}
		\label{fig:salientann}
	\end{subfigure}	
	\begin{subfigure}[t]{0.2\linewidth}
		\centering
		\includegraphics[width=0.95\linewidth,keepaspectratio]{images/user_study/000000013729.jpg}
		\caption{\textbf{Original}}
		\label{fig:userstudy-i}
	\end{subfigure}%
	\begin{subfigure}[t]{0.2\linewidth}
		\centering
		\includegraphics[width=0.95\linewidth,keepaspectratio]{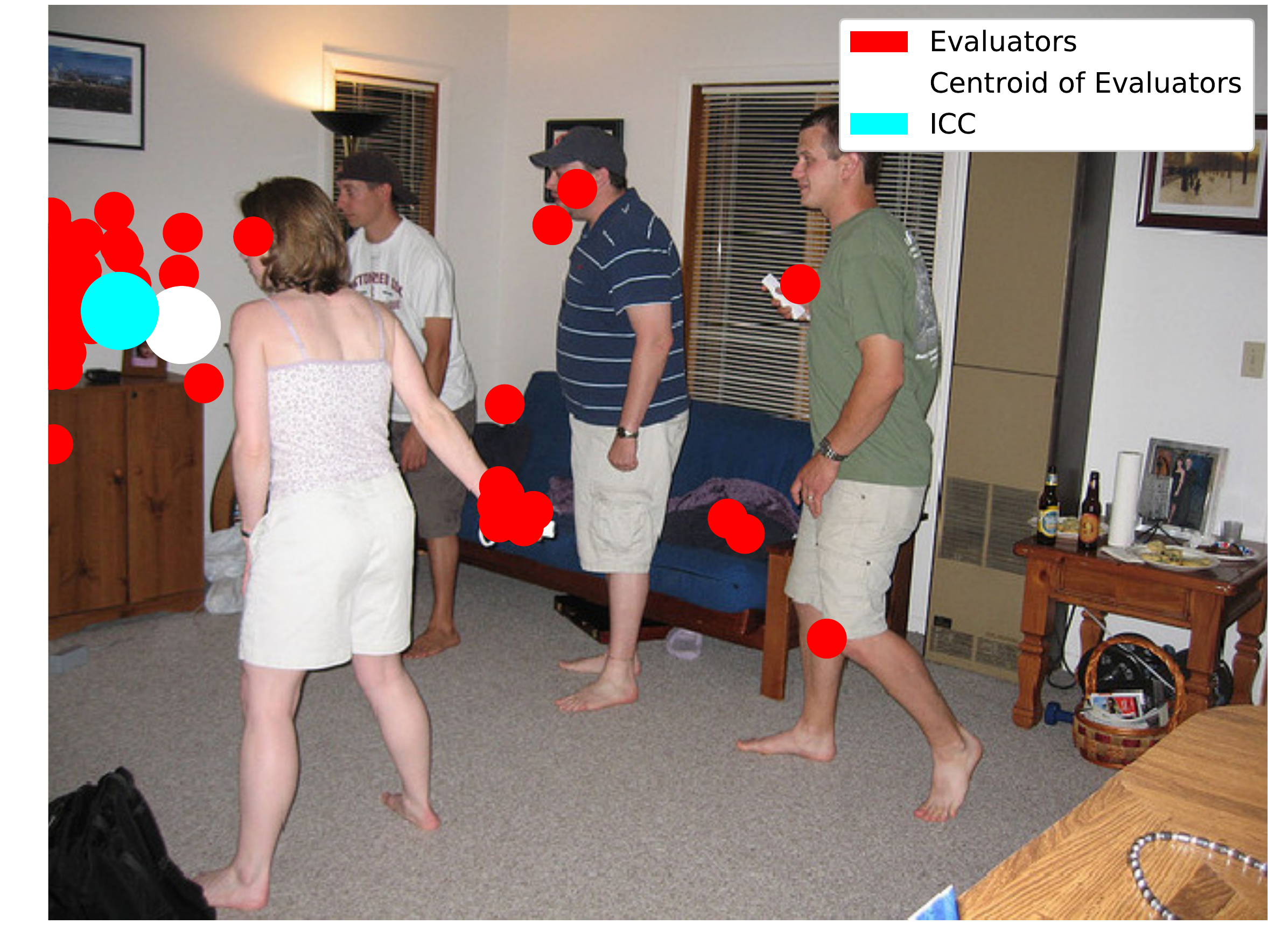}
		\caption{\textbf{AR}}
		\label{fig:userstudy-j}
	\end{subfigure}%
	\begin{subfigure}[t]{0.2\linewidth}
		\centering
		\includegraphics[width=0.95\linewidth, keepaspectratio]{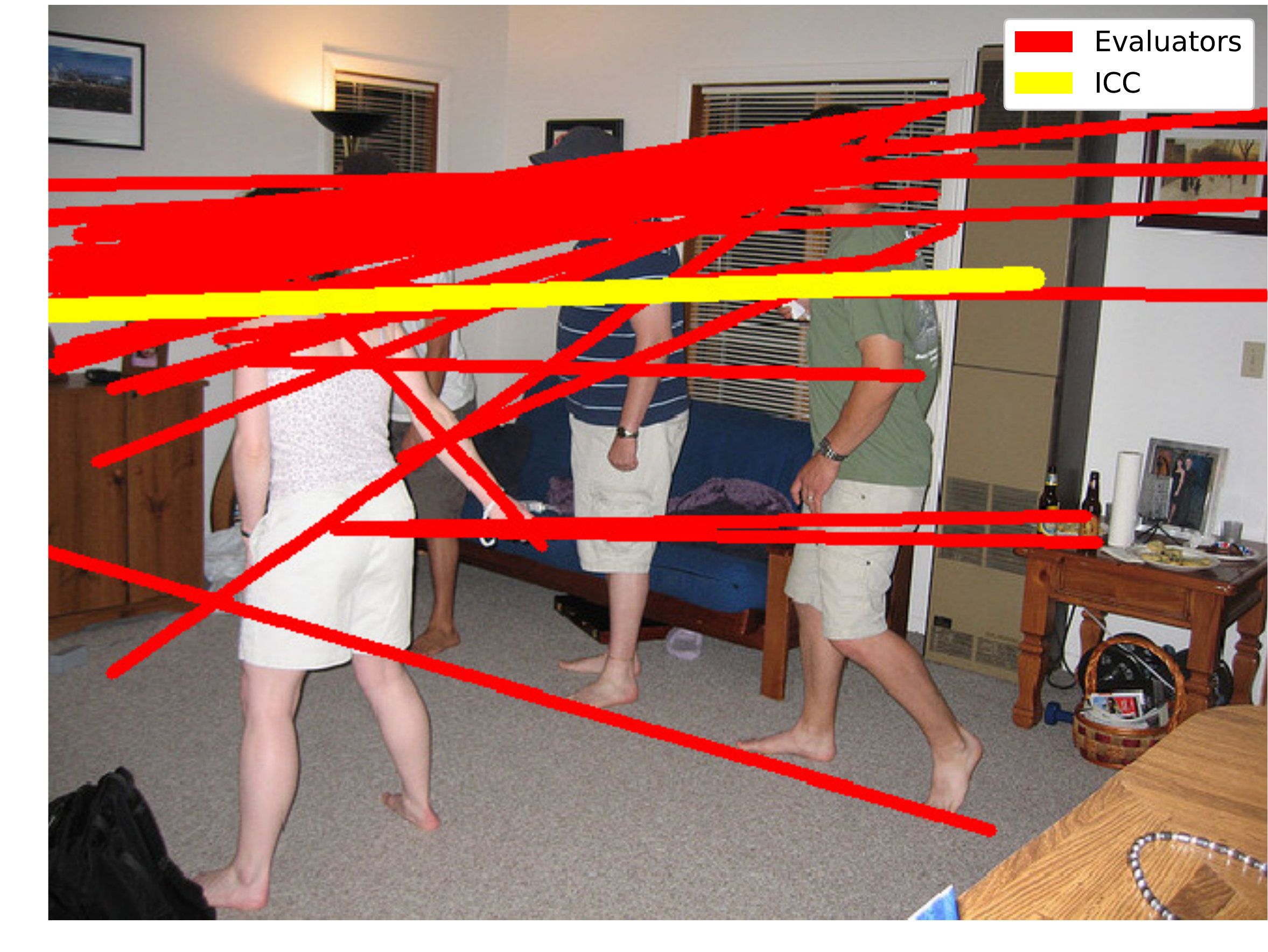}
		\caption{\textbf{AL}}
		\label{fig:userstudy-k}
	\end{subfigure}%
	\begin{subfigure}[t]{0.2\linewidth}
		\centering
		\includegraphics[width=0.95\linewidth, keepaspectratio]{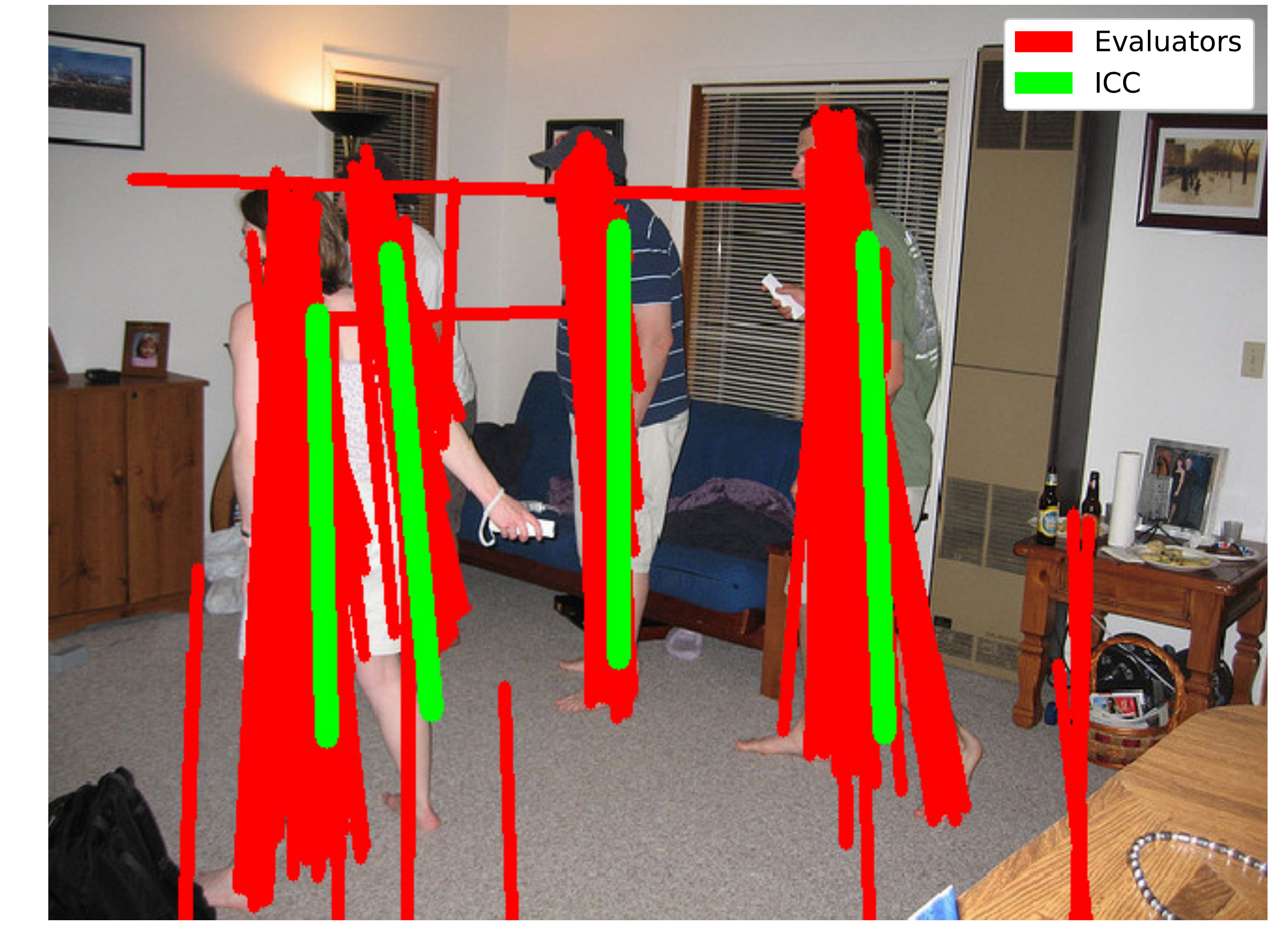}
		\caption{\textbf{PL}}
		\label{fig:userstudy-l}
	\end{subfigure}%
	\begin{subfigure}[t]{0.2\linewidth}
		\centering
		\includegraphics[width=0.95\linewidth,keepaspectratio]{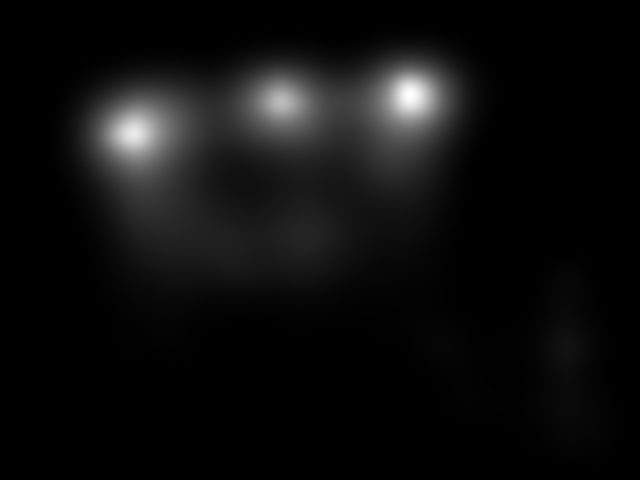}
		\caption{\textbf{Eye-Fix$^n$}}
		\label{fig:salientcoco}
	\end{subfigure}	
\caption{User Study analysis: \nth{1} row shows an example of \emph{The kiss of Judas}, \nth{2} of \emph{annunciation} and \nth{3} of \emph{COCO}\protect{\cite{linMicrosoftCOCOCommon2015}}. \protect{\subref{fig:userstudy-i}} column are original images; \protect{\subref{fig:userstudy-j}} shows the Action Region (AR) by ICC (cyan), all Evaluators (red) and the centroid (white) of all annotators; \protect{\subref{fig:userstudy-k}} and \protect{\subref{fig:userstudy-l}} show AL and PL by ICC (yellow, green) and all annotators (red, red) respectively; and \protect{\subref{fig:salientcoco}} highlights the eye-fixation regions~\protect{\cite{cornia2018predicting}}}
\label{fig:userstudy}
\end{figure}
	
\subsubsection{Evaluation Metrics.} We evaluate each of AL, AR and PL individually and for all the annotators. We compare the performances between: E \emph{vs}.\ ICC, NE \emph{vs}.\ ICC and E \emph{vs}.\ NE. 
The standard deviation ($SD_{AR}$) in the labeling of ARs is a measure of agreement between the annotators, lower $SD_{AR}$ means they agree more where as higher $SD_{AR}$ means they disagree more. For comparison of labeled ARs and predicted ARs, the $L_{2}$ distance is considered between E/NE and ICC. 
For AL and PL, the lines are considered as sets of points in the image. We find the distance between these two sets using Hausdorff Distance (HD)~\cite{hd}. We also calculate the angular deviation (AD) between the AL of the E/NE and the ICC. Higher angular deviation means higher value of AD. 
    
\subsection{Results and Discussion}\label{ExRes}
\subsubsection{Action Lines.}
Each AL represents an important action or geometry in the original painting, however, both differ in their presentation (\cf Max Imdahl's~\emph{et al.} manually created analysis in \cref{fig:main-f}). It is not easy to quantify such expressions since they are more visually interpretive than quantitatively. In \nth{1} row of \cref{fig:main-e}, Giotto's ``Kiss of Judas'', Judas is hugging Jesus. Many people are directly looking at them. This structure is apparent in other works of Giotto. The average slope of all these gazes creates the AL (yellow line in \cref{fig:main-g}). This AL gives us information about the direction from which these important points have been viewed. Sometimes, however, OpenPose fails to detect human poses, especially the poses of Jesus and Judas were not recognized (due to occlusions), which could have had a strong impact on the AL. Giotto's paintings sometimes also have a line of heads, with one head next to each other and most of them on the same height in the image (\cref{fig:main-e}). In this case, we see that our method does a very good job of detecting the AL (\cref{fig:main-g}).
\subsubsection{Action Regions.} The gaze structure in Giotto's paintings also motivates us to use their gaze directions in finding the approximate region of action. Since the gaze directions are conditioned on the body keypoints (\cref{sec:GazeEstCorr}.b) they also have pose context. We use the intersection of all gazes to identify important regions in the picture. In \cref{fig:main-g}, we see the locations of the ARs are very interesting. These regions are the center of high activity, which follows our definition of ARs~(\cref{sec:actions}). We also observe that generally the AL passes through ARs. \cref{fig:circlemap-b,fig:circlemap-c,fig:salientfuss} show an interesting case with multiple ARs and ALs. Our ICC is capable of predicting 2 ALs and their corresponding ARs. We see that some people focus on the 2 central characters accounting for the \nth{1} AL, while the 2 central characters are looking at the sitting character creating the \nth{2} AL. The corresponding eye-fixation map in \cref{fig:salientfuss} highlights the regions where the 2 ALs are passing through and where the 2 ARs are located. 
\subsubsection{Semantic Foreground/Background separation.}
The main characters usually form the foreground of the image. Additionally, using their poses, we detect which of the areas in the image constitute foreground/ background. We in-paint the image at their body positions. In order to achieve this, K-means clustering is applied with a small k-value (good clustering and a filled, crack-less foreground surface). The small k-value could lead, in some cases, to a completely incorrect separation if the same colors were present in the foreground/background. Giotto's paintings also show the geometry of foreground/ background separation (\cf~\cref{fig:main-f}) to be an important underlying structuring element. This separation is apparent in \cref{fig:main-g,fig:pose-c} as the big light brown region covering half of the image and the remaining blue backdrop as the background.
    \begin{figure}[t]
		\centering
		\begin{subfigure}[t]{0.25\linewidth}
			\centering
			\includegraphics[width=0.9\linewidth,height=0.8\linewidth]{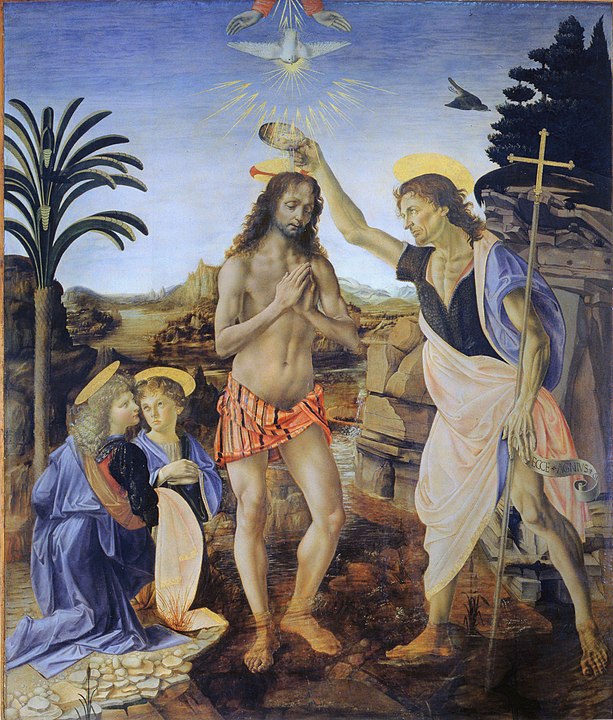}
			\label{fig:crossdom-a}
		\end{subfigure}%
		\begin{subfigure}[t]{0.25\linewidth}
			\centering
			\includegraphics[width=0.9\linewidth,height=0.8\linewidth]{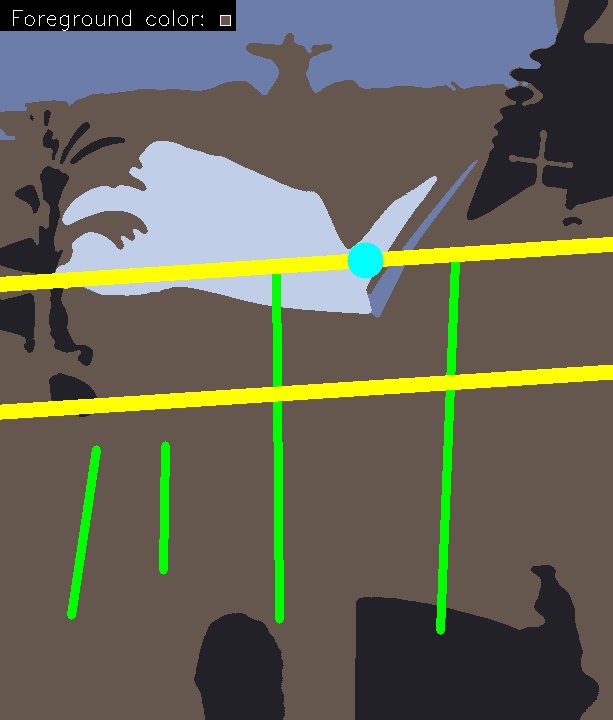}
			\label{fig:crossdom-b}
		\end{subfigure}%
		\begin{subfigure}[t]{0.25\linewidth}
			\centering
			\includegraphics[width=0.9\linewidth,height=0.8\linewidth]{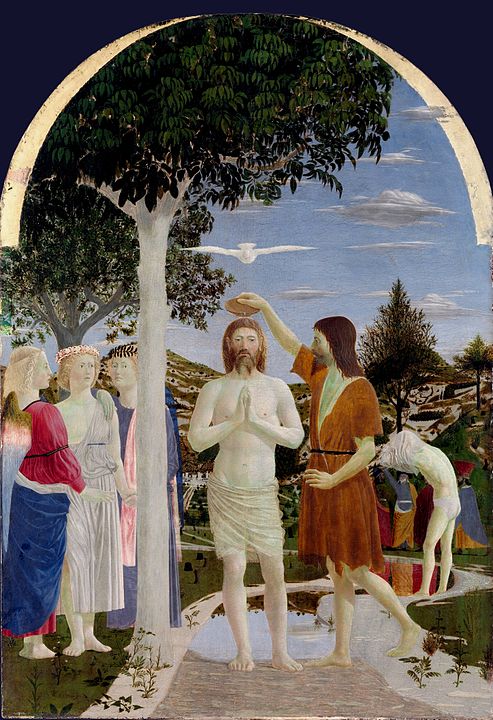}
			\label{fig:crossdom-c}
		\end{subfigure}%
		\begin{subfigure}[t]{0.25\linewidth}
			\centering
			\includegraphics[width=0.9\linewidth,height=0.8\linewidth]{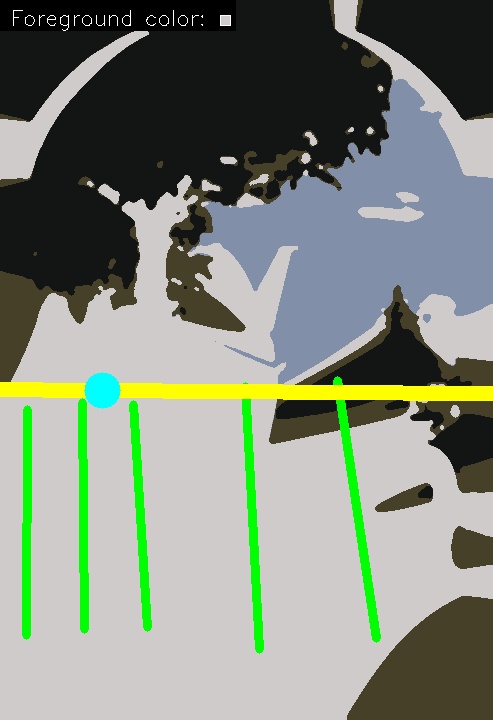}
			\label{fig:crossdom-d}
		\end{subfigure}
		\begin{subfigure}[t]{0.25\linewidth}
			\centering
			\includegraphics[width=0.9\linewidth,height=2cm]{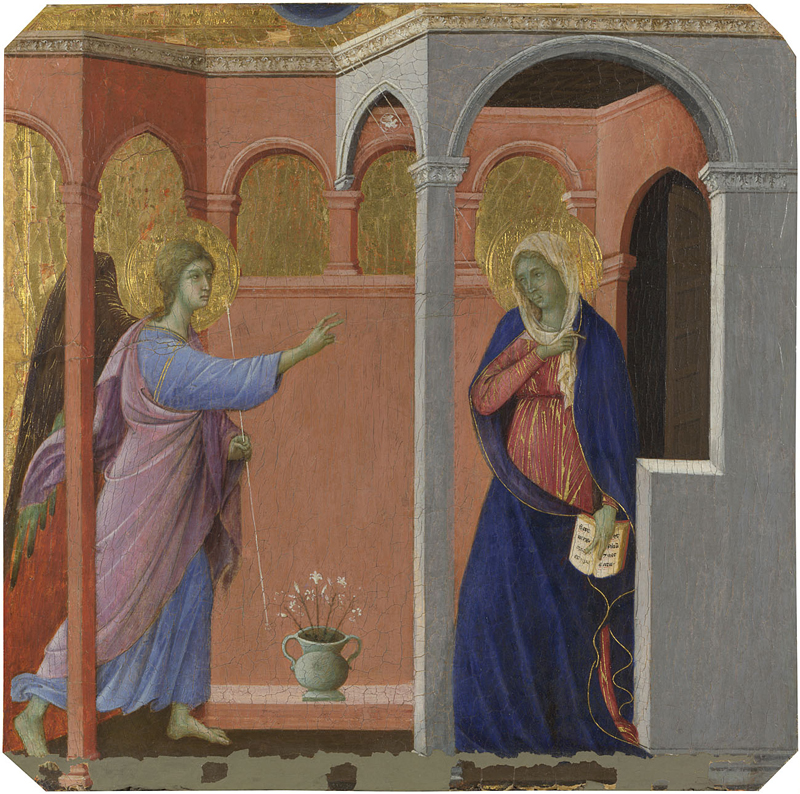}
			\label{fig:crossdom-e}
		\end{subfigure}%
		\begin{subfigure}[t]{0.25\linewidth}
			\centering
			\includegraphics[width=0.9\linewidth,height=2cm]{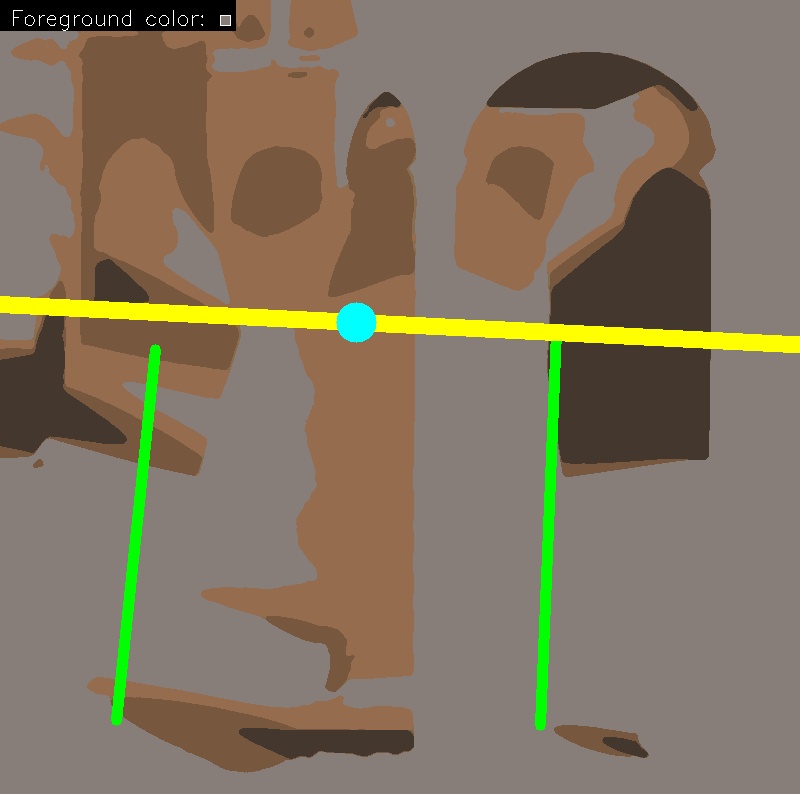}
			\label{fig:crossdom-f}
		\end{subfigure}%
		\begin{subfigure}[t]{0.25\linewidth}
			\centering
			\includegraphics[width=0.9\linewidth,height=2cm]{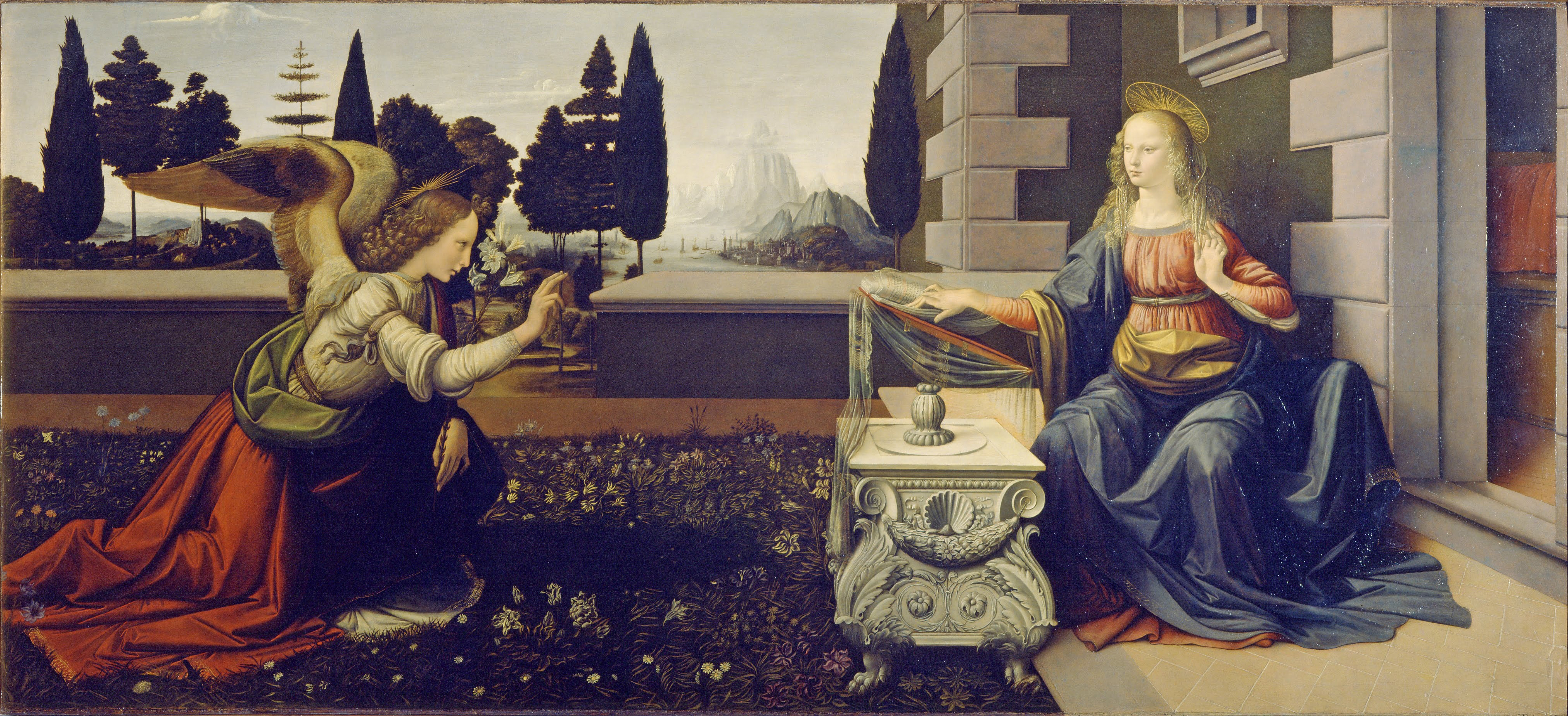}
			\label{fig:crossdom-g}
		\end{subfigure}%
		\begin{subfigure}[t]{0.25\linewidth}
			\centering
			\includegraphics[width=0.9\linewidth,height=2cm]{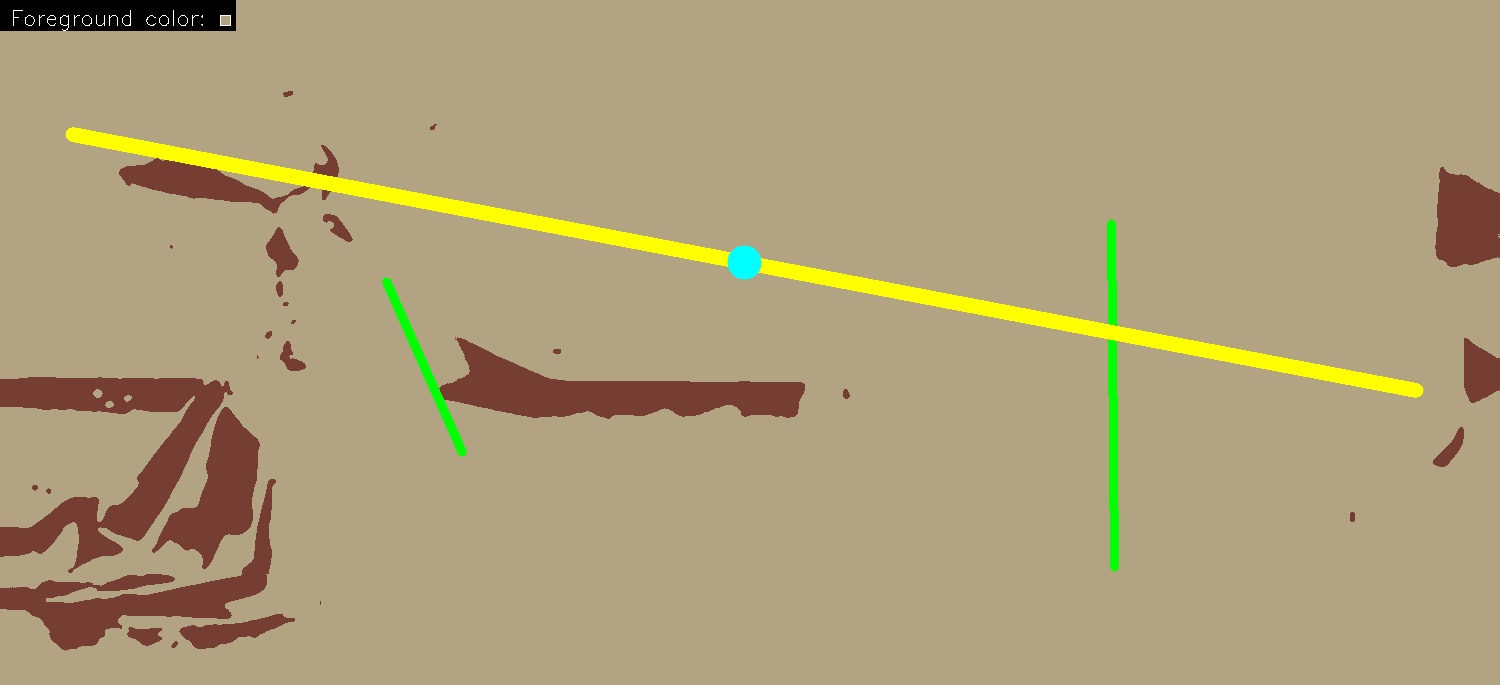}
			\label{fig:crossdom-h}
		\end{subfigure}
		\begin{subfigure}[t]{0.25\linewidth}
			\centering
			\includegraphics[width=0.9\linewidth,keepaspectratio]{crossdomain/9590.jpg}
			\caption{\textbf{Original}}
			\label{fig:crossdom-i}
		\end{subfigure}%
		\begin{subfigure}[t]{0.25\linewidth}
			\centering
			\includegraphics[width=0.9\linewidth,keepaspectratio]{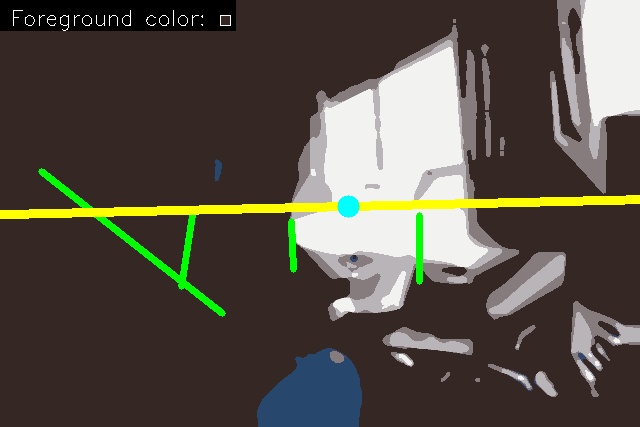}
			\caption{\textbf{ICC}}
			\label{fig:crossdom-j}
		\end{subfigure}%
		\begin{subfigure}[t]{0.25\linewidth}
			\centering
			\includegraphics[width=0.9\linewidth, keepaspectratio]{crossdomain/5193.jpg}
			\caption{\textbf{Original}}
			\label{fig:crossdom-k}
		\end{subfigure}%
		\begin{subfigure}[t]{0.25\linewidth}
			\centering
			\includegraphics[width=0.9\linewidth, keepaspectratio]{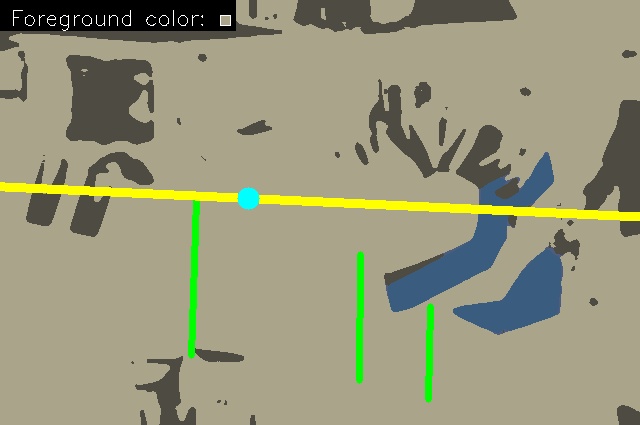}
			\caption{\textbf{ICC}}
			\label{fig:crossdom-l}
		\end{subfigure}
    \caption{Cross domain analysis: \nth{1} row depicts \emph{Baptism}, \nth{2} shows \emph{annunciation}, and the \nth{3} are images from \emph{COCO}~\protect{\cite{linMicrosoftCOCOCommon2015}}. \protect{\subref{fig:crossdom-i}} and \protect{\subref{fig:crossdom-k}} columns are original images, \protect{\subref{fig:crossdom-j}} and \protect{\subref{fig:crossdom-l}} are their corresponding \emph{ICC}.}
    \label{fig:crossdom}
\end{figure}
	
\subsection{Cross-Domain Adaptability}\label{crossdom}
The first $2$ source (\cref{fig:crossdom},~\nth{1}~\&~\nth{2}~row), images are taken from $2$ different iconographies: \emph{Baptism} and \emph{Annunciation}. For the third, (\cref{fig:crossdom}, \nth{3}~row), we use images from COCO -- a dataset of images from everyday life. We choose images that contains people doing different activities.

\cref{fig:crossdom-i,fig:crossdom-j} depict examples where our method is able to locate the AR, AL and PL very effectively. In addition, the foreground/background separation is also clearly visible in the ICC. For example in \nth{2} row of \cref{fig:crossdom-j}, we see that the PLs are correct, the AL passes through the main activity and the AR is localized between the two characters. In \nth{1} row of \cref{fig:crossdom-j}, our method is able to detect the presence of 2 ALs: one aligns with the gaze of the female and the waist of the male, and the other aligns through the heads of the males. The AR is also very well localized, with good foreground/background separation. 

In \cref{fig:crossdom-k,fig:crossdom-l}, our method fails either for ALs, or the foreground/ background separation. However, sometimes the ARs are very well localized even in these examples. In \nth{1} row of \cref{fig:crossdom-l}, we can observe that the main action region and background-foreground is incorrect, while the AL seems to be acceptable. Similarly, in \nth{2} and \nth{3} rows of \cref{fig:crossdom-l}, the foreground/background separation is not good and neither is the AL, but the AR is localized very well between the protagonists.

\begin{table}[t]
\centering
\caption{Quantitative Evaluation. \nth{1} and \nth{2} columns show the Standard Deviation ($SD_{AR}$) for AR amongst Experts (E) and Non-Experts (NE). \nth{3}, \nth{4} and \nth{5} columns show the $L_{2}$ distance between centroids of E/ICC, NE/ICC and E/NE. \nth{6} column shows the HD between all annotators (ALL) and ICC. \nth{7} column shows cosine angular deviation (degrees) between ALL/ICC.}
\label{tab:quanteval}
\begin{tabular}{cc c@{\hspace{0.7em}}ccc c@{\hspace{0.7em}}c c@{\hspace{0.7em}}c}
\toprule
\multicolumn{2}{c}{$\mathbf{SD_{AR}}$} && \multicolumn{3}{c}{$\mathbf{L_{2}}$} && \textbf{HD} && \textbf{AD} \\
\cmidrule(lr){1-2}\cmidrule(lr){4-6}\cmidrule(lr){8-9}\cmidrule(lr){9-10}
\textbf{E} & \textbf{NE} && \textbf{E/ICC} $|$ & \textbf{NE/ICC} $|$ & \textbf{E/NE} && \textbf{ALL/ICC} && \textbf{ALL/ICC} \\ 
\midrule
$0.054$ & $0.089$ && $0.187$ & $0.180$ & $0.035$ && $664.35$ && $\ang{36.57}$ \\ \bottomrule
\end{tabular}
\end{table}

\subsection{Quantitative Evaluation of User Study}\label{quanteval}
For comparing action regions (AR) we reduce the image sizes to $1\times1$. From \cref{tab:quanteval}, we can see that the Es have more agreement among-st themselves for ARs as compared to NEs. However, the Euclidean distance ($L_{2}$) between the centroid of the Es and the ICC (E/ICC) has a similar value to that of the NEs and the ICC (NE/ICC), indicating that the region of ARs has high agreement for all groups, \ie E, NE and ICC (people \textit{vs.} \emph{ICC}). The low value of $L_{2}$ between E/NE shows that they concur about the ARs (an error of $3.5\%$ for the localization of AR). This fact is also recognizable in \cref{fig:userstudy-j}, where the centroids of all annotators lie very close to that proposed by our ICC. The eye (or gaze)-fixation maps of these images are also shown in \cref{fig:salientcoco}. These maps show those regions of the image where the gaze of an observer would focus. These maps are complementary to the compositional elements like ALs and ARs. The structure of these maps follows the slope of ALs and are distributed along the AL and the AR.   

HD between all the annotators and ICC (ALL/ICC) is quite low (\SI{38}{\percent}) than the worst HD distance (ALL/Infinity: $1070.2$ for the case of end points of the diagonal) showing that the ALs of our ICC has very good correlation to all annotators. We also see that the angular distance (AD) of all ALs with that of our ICC has an average value of $36.57^\circ$ (ALL/ICC).  

We observed that there is a positive correlation (0.2) between the images with low $SD_{AR}$ and their corresponding $L_{2(AR)}$,
meaning that when the annotators’ agreement for the position of AR was higher, our
method predicted AR closer to the labeled ones.
Similarly, we observed opposite trend for high $SD_{AR}$ images, with correlation of $-0.790$, meaning that when
there is higher disagreement between the annotators about the location of the AR,
our method predicted AR farther away.
Similar trend is observed when we correlate $SD_{AR}$ with $HD$. When annotators agree (low $SD_{AR}$), the predicted AL is closer to the labeled ones, where as when annotators disagree (high $SD_{AR}$), the predicted AL is farther to the labeled one. 
	
\section{Conclusion and Future Work}\label{Conc}
With the help of existing machine learning tools, state-of-the-art pose estimation framework and image analysis, we have presented a novel no-training approach to understand underlying structures in art historical scenes. We show that recognizing key elements within a scene, such as ALs, ARs and FG/BG separation, helps in understanding the composition of any scene. Apart from using a pre-trained OpenPose framework, our method does not need any training. This makes our method very light, robust and applicable to any scene for analysis, especially when there is no ground truth available, which is often the case in art historic images. It can be used to pre-compare images for image retrieval and analysis. We also showed that our method works extremely well for images from related domains as well as images from everyday life. 

Our method is a complementary approach to the perception of semantics in artworks~\cite{garcia2018read} and iconography in general. It is one of the first attempts at understanding scenes or paintings in art history grounded in compositional elements. Future development could include training OpenPose on art historical data, and including the scene information to explore the role of scene narratives on the underlying compositions. At the same time our algorithm can be applied to various target domains where composition lines and narratives are used.  

\clearpage
%
%
\bibliographystyle{splncs04}
\bibliography{0014}
\end{document}